\documentclass[runningheads]{llncs}
\usepackage{graphicx}
\usepackage{subcaption}
\usepackage{comment}
\usepackage{amsmath,amssymb} 
\usepackage{color}
\usepackage{dsfont}
\usepackage{booktabs}
\usepackage{xcolor}
\usepackage{colortbl}
\usepackage{multirow}
\usepackage{paralist}
\usepackage{marvosym}
\usepackage[export]{adjustbox}
\usepackage[pagebackref=true,breaklinks=true,letterpaper=true,colorlinks,bookmarks=false,citecolor=blue,linkcolor=blue]{hyperref}
\usepackage{makecell}

\newcommand{\ch}{\checkmark}
\definecolor{sh_gray}{rgb}{0.84,0.84,0.84}
\definecolor{sh_gray2}{rgb}{1,0.89,0.75}
\definecolor{color3}{rgb}{0.95,0.95,0.95}
\definecolor{color4}{rgb}{0.96,0.96,0.86}
\definecolor{color5}{rgb}{0.90,0.90,0.90}

\newcolumntype{M}[1]{>{\centering\arraybackslash}m{#1}}

\usepackage[width=122mm,left=12mm,paperwidth=146mm,height=193mm,top=12mm,paperheight=217mm]{geometry}

\begin{document}
\pagestyle{headings}
\mainmatter

\title{Learning Enriched Features for Real Image Restoration and Enhancement}


\titlerunning{Learning Enriched Features for Real Image Restoration and Enhancement}
%
\author{
Syed Waqas Zamir\inst{1} \and
Aditya Arora\inst{1} \and
Salman Khan\inst{1,2} \and
Munawar Hayat\inst{1,2} \and \\
Fahad Shahbaz Khan\inst{1,2} \and
Ming-Hsuan Yang\inst{3,4} \and
Ling Shao\inst{1,2}
}
\authorrunning{Zamir et al.}
%
\institute{Inception Institute of Artificial Intelligence, UAE\and
Mohamed bin Zayed University of Artificial Intelligence, UAE\and
University of California, Merced, USA \and
Google Research}
\maketitle

\vspace{-0.5em}
\begin{abstract}
With the goal of recovering high-quality image content from its degraded version, image restoration enjoys numerous applications, such as in surveillance, computational photography, medical imaging, 
and remote sensing. 
Recently, convolutional neural networks (CNNs) have achieved dramatic improvements over conventional approaches for image restoration task.
Existing CNN-based methods typically operate either on full-resolution or on progressively low-resolution representations. In the former case, spatially precise but contextually less robust results are achieved, while in the latter case, semantically reliable but spatially less accurate outputs are generated.
In this paper, we present a novel architecture with the collective goals of maintaining spatially-precise high-resolution representations through the entire network, and receiving strong contextual information from the low-resolution representations. 
The core of our approach is a multi-scale residual block containing several key elements: (a) parallel multi-resolution convolution streams for extracting multi-scale features, (b) information exchange across the multi-resolution streams, (c) spatial and channel attention mechanisms for capturing contextual information, and (d) attention based multi-scale feature aggregation.   
In a nutshell, our approach learns an enriched set of features that combines contextual information from multiple scales, while simultaneously preserving the high-resolution spatial details.
Extensive experiments on five real image benchmark datasets demonstrate that our method, named as MIRNet, achieves state-of-the-art results for a variety of image processing tasks, including image denoising, super-resolution and image enhancement. 
The source code and pre-trained models are available at \url{https://github.com/swz30/MIRNet}.

\vspace{-0.5em}
\keywords{Image denoising, super-resolution, and image enhancement. }
\end{abstract}

\section{Introduction}
Image content is exponentially growing due to the ubiquitous presence of cameras on various devices.  
During image acquisition, degradations of different severity are often introduced.
It is either because of the physical limitations of cameras or due to inappropriate lighting conditions.
For instance, smartphone cameras come with a narrow aperture and have small sensors with limited dynamic range. 
Consequently, they frequently generate noisy and low-contrast images. 
Similarly, images captured under the unsuitable lighting are either too dark or too bright.
The art of recovering the original clean image from its corrupted measurements is studied under the image restoration task. It is an ill-posed inverse problem, due to the existence of many possible solutions.

Recently, deep learning models have made significant advancements for image restoration and enhancement, as they can learn strong (generalizable) priors from large-scale datasets. 
Existing CNNs typically follow one of the two architecture designs: 1) an encoder-decoder, or 2) high-resolution (single-scale) feature processing. 
The encoder-decoder models \cite{ronneberger2015u,kupyn2019deblurgan,chen2018,zhang2019kindling} first progressively map the input to a low-resolution representation, and then apply a gradual reverse mapping to the original resolution. Although these approaches learn a broad context by spatial-resolution reduction, on the downside, the fine spatial details are lost, making it extremely hard to recover them in the later stages. On the other side, the high-resolution (single-scale) networks \cite{dong2015image,DnCNN,zhang2020residual,ignatov2017dslr} do not employ any downsampling operation, and thereby produce images with spatially more accurate details. However, these networks are less effective in encoding contextual information due to their limited receptive field.

Image restoration is a position-sensitive procedure, where pixel-to-pixel correspondence from the input image to the output image is needed. Therefore, it is important to remove only the undesired degraded image content, while carefully preserving the desired fine spatial details (such as true edges and texture).
Such functionality for segregating the degraded content from the true signal can be better incorporated into CNNs with the help of large context, \textit{e.g.}, by enlarging the receptive field. 
Towards this goal, we develop a new \emph{multi-scale} approach that maintains the original high-resolution features along the network hierarchy, thus minimizing the loss of precise spatial details. Simultaneously, our model encodes multi-scale context by using \emph{parallel convolution streams} that process features at lower spatial resolutions. 
The multi-resolution parallel branches operate in a manner that is complementary to the main high-resolution branch, thereby providing us more precise and contextually enriched feature representations. 

The main difference between our method and existing multi-scale image processing approaches is the way we aggregate contextual information. 
First, the existing methods \cite{tao2018scale,nah2017,gu2019self} process each scale in isolation, and exchange information only in a  top-down manner. In contrast, we progressively fuse information across all the scales at each resolution-level, allowing both top-down and bottom-up information exchange. 
Simultaneously, both fine-to-coarse and coarse-to-fine knowledge exchange is laterally performed on each stream by a new \emph{selective kernel} fusion mechanism. Different from existing methods that employ a simple concatenation or averaging of features coming from multi-resolution branches, our fusion approach dynamically selects the useful set of kernels from each branch representations using a self-attention approach. 
More importantly, the proposed fusion block combines features with varying receptive fields, while preserving their distinctive complementary characteristics.

\noindent Our main contributions in this work include:\vspace{-0.5em}
\begin{itemize}
\item A novel feature extraction model that obtains a complementary set of features across multiple spatial scales, while maintaining the original high-resolution features to preserve precise spatial details. 
\item A regularly repeated mechanism for information exchange, where the features across multi-resolution branches are progressively fused together for improved representation learning.
\item A new approach to fuse multi-scale features using a selective kernel network that dynamically combines variable receptive fields and faithfully preserves the original feature information at each spatial resolution. 
\item A recursive residual design that progressively breaks down the input signal in order to simplify the overall learning process, and allows the construction of very deep networks. 
\item Comprehensive experiments are performed on five real image benchmark datasets for different image processing tasks including, image denoising, super-resolution and image enhancement. Our method achieves state-of-the-results on \textit{all} five datasets. Furthermore, we extensively evaluate our approach on practical challenges, such as generalization ability across datasets.
\end{itemize}

\section{Related Work}
With the rapidly growing image content, there is a pressing need to develop effective image restoration and enhancement algorithms. 
In this paper, we propose a new method capable of performing image denoising, super-resolution and image enhancement. Unlike existing works for these problems, our approach processes features at the original resolution in order to preserve spatial details, while effectively fuses contextual information from multiple parallel branches. Next, we briefly describe the representative methods for each of the studied problems. 
%

\vspace{0.4em}\noindent\textbf{Image denoising.} 
Classic denoising methods are mainly based on modifying transform coefficients \cite{yaroslavsky1996local,donoho1995noising,simoncelli1996noise} or averaging neighborhood pixels \cite{smith1997susan,tomasi1998bilateral,perona1990scale,rudin1992nonlinear}. 
Although the classical methods perform well, the self-similarity \cite{efros1999texture} based algorithms, \textit{e.g.}, NLM \cite{NLM} and BM3D \cite{BM3D}, demonstrate promising denoising performance. 
Numerous patch-based algorithms that exploit redundancy (self-similarity) in images are later developed \cite{dong2012nonlocal,WNNM,mairal2009non,hedjam2009markovian}. 
Recently, deep learning-based approaches \cite{MLP,RIDNet,Brooks2019,Gharbi2016,CBDNet,N3Net,DnCNN,FFDNetPlus,Zamir2020CycleISP} make significant advances in image denoising, yielding favorable results than those of the hand-crafted methods.

%
\vspace{0.4em} \noindent\textbf{Super-resolution (SR).} 
Prior to the deep-learning era, numerous SR algorithms have been proposed based on the sampling theory~\cite{keys1981cubic,irani1991improving}, edge-guided interpolation \cite{allebach1996edge,zhang2006edge}, natural image priors \cite{kim2010single,xiong2010robust}, patch-exemplars \cite{chang2004super,freedman2011image} and sparse representations \cite{yang2010image,yang2008image}. 
Currently, deep-learning techniques are actively being explored, as they provide dramatically improved results over conventional algorithms. 
The data-driven SR approaches differ according to their architecture designs~\cite{wang2019deep,anwar2019deep,ntire2019_superresolution}. Early methods~\cite{dong2014learning,dong2015image} take a low-resolution (LR) image as input and learn to directly generate its high-resolution (HR) version. 
In contrast to directly producing a latent HR image, recent SR networks \cite{VDSR,tai2017memnet,tai2017image,hui2018fast} employ the residual learning framework \cite{He2016} to learn the high-frequency image detail, which is later added to the input LR image to produce the final super-resolved result.
Other networks designed to perform SR include recursive learning \cite{kim2016deeply,han2018image,ahn2018fast}, progressive reconstruction \cite{wang2015deep,Lai2017}, dense connections \cite{tong2017image,wang2018esrgan,zhang2020residual}, attention mechanisms \cite{RCAN,dai2019second,zhang2019residual}, multi-branch learning \cite{Lai2017,EDSR,dahl2017pixel,li2018multi}, and generative adversarial networks (GANs) \cite{wang2018esrgan,park2018srfeat,sajjadi2017enhancenet,SRResNet}. 
%
%

\vspace{0.4em}\noindent\textbf{Image enhancement.}
Oftentimes, cameras generate images that are less vivid and lack contrast. A number of factors contribute to the low quality of images, including unsuitable lighting conditions and physical limitations of camera devices. 
For image enhancement, histogram equalization is the most commonly used approach. However, it frequently produces under- or over-enhanced images. Motivated by the Retinex theory \cite{land1977retinex}, several enhancement algorithms mimicking human vision have been proposed in the literature \cite{bertalmio2007,palma2008perceptually,jobson1997multiscale,rizzi2004retinex}.
Recently, CNNs have been successfully applied to general, as well as low-light, image enhancement problems \cite{ntire2019_enhancement}. Notable works employ Retinex-inspired networks \cite{Shen2017,wei2018deep,zhang2019kindling}, encoder-decoder networks \cite{chen2018encoder,Lore2017,ren2019low}, and GANs \cite{chen2018deep,ignatov2018wespe,deng2018aesthetic}.
%


\begin{figure}[t]
\begin{center}
\begin{tabular}[t]{c} \hspace{-2mm}
\includegraphics[width=\textwidth]{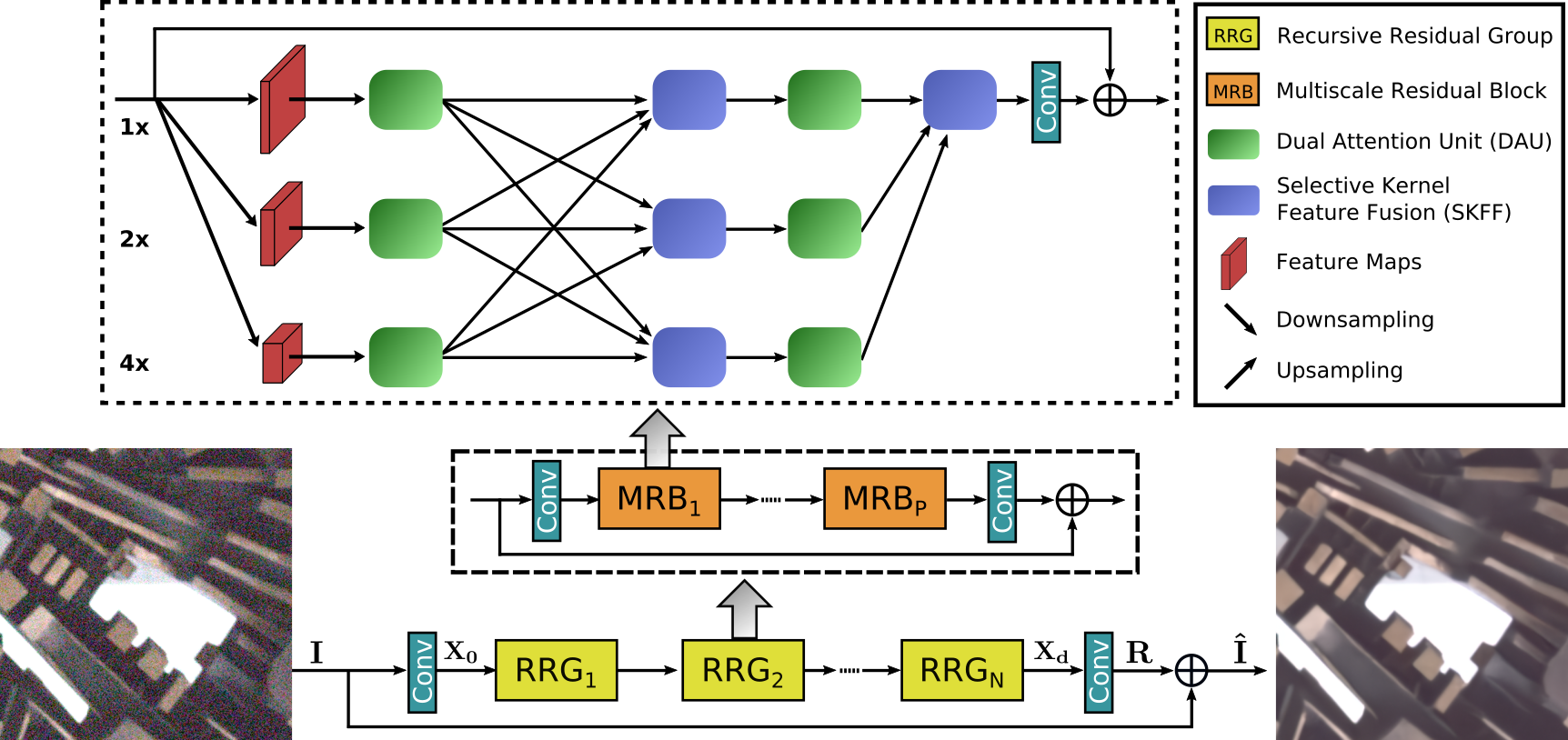}
\end{tabular}
\end{center}
\vspace*{-4mm}
\caption{\small Framework of the proposed network MIRNet that learns enriched feature representations for image restoration and enhancement. MIRNet is based on a recursive residual design. In the core of MIRNet is the multi-scale residual block (MRB) whose main branch is dedicated to maintaining spatially-precise high-resolution representations through the entire  network and the complimentary set of parallel branches provide better contextualized features. It also allows information exchange across parallel streams via selective kernel feature fusion (SKFF) in order to consolidate the high-resolution features with the help of low-resolution features, and vice versa.}
\label{fig:framework}
\vspace{-1em}
\end{figure}

\section{Proposed Method}
In this section, we first present an overview of the proposed MIRNet for image restoration and enhancement, illustrated in Fig.~\ref{fig:framework}.
We then provide details of the \emph{multi-scale residual block}, which is the fundamental building block of our method, containing several key elements: \textbf{(a)} parallel multi-resolution convolution streams for extracting (fine-to-coarse) semantically-richer and (coarse-to-fine) spatially-precise feature representations, \textbf{(b)} information exchange across multi-resolution streams, \textbf{(c)} attention-based aggregation of features arriving from multiple streams, \textbf{(d)} dual-attention units to capture contextual information in both spatial and channel dimensions, and \textbf{(e)} residual resizing modules to perform downsampling and upsampling operations.

\vspace{0.4em} \noindent \textbf{Overall Pipeline.} 
Given an image $\mathbf{I} \in \mathbb{R}^{H\times W \times 3}$, the network first applies a convolutional layer to extract low-level features $\mathbf{X_0} \in \mathbb{R}^{H\times W \times C}$.
Next, the feature maps $\mathbf{X_0}$ pass through $N$ number of recursive residual groups (RRGs), yielding deep features $\mathbf{X_d} \in \mathbb{R}^{H\times W \times C}$. 
We note that each RRG contains several multi-scale residual blocks, which is described in Section~\ref{sec:msrb}. 
Next, we apply a convolution layer to deep features $\mathbf{X_d}$ and obtain a residual image $\mathbf{R} \in \mathbb{R}^{H\times W \times 3}$. 
Finally, the restored image is obtained as $\mathbf{\hat{I}} = \mathbf{I} + \mathbf{R}$.   
We optimize the proposed network using the Charbonnier loss \cite{charbonnier1994}:

\begin{equation}
\label{Eq:loss}
\mathcal{L}(\mathbf{\hat{I}},\mathbf{I}^*) = \sqrt{ {\|\mathbf{\hat{I}}-\mathbf{I}^*\|}^2 + {\varepsilon}^2 },
\end{equation}
where $\mathbf{I}^*$ denotes the ground-truth image, and $\varepsilon$ is a constant which we empirically set to $10^{-3}$ for all the experiments.
%

\subsection{Multi-scale Residual Block (MRB)}
\label{sec:msrb}
In order to encode context, existing CNNs \cite{ronneberger2015u,newell2016stacked,noh2015learning,xiao2018simple,badrinarayanan2017segnet,peng2016recurrent} typically employ the following architecture design: \textbf{(a)} the receptive field of neurons is fixed in \textit{each} layer/stage, \textbf{(b)} the spatial size of feature maps is \textit{gradually} reduced to generate a semantically strong low-resolution representation, and \textbf{(c)} a high-resolution representation is \textit{gradually} recovered from the low-resolution representation.  
However, it is well-understood in vision science that in the primate visual cortex, the sizes of the local receptive fields of neurons in the same region are different \cite{hubel1962receptive,riesenhuber1999hierarchical,serre2007robust,hung2005fast}. 
Therefore, such a mechanism of collecting multi-scale spatial information in the same layer needs to be incorporated in CNNs~\cite{huang2017multi,hrnet,fourure2017residual,Szegedy2015}.
In this paper, we propose the multi-scale residual block (MRB), as shown in Fig.~\ref{fig:framework}. 
It is capable of generating a spatially-precise output by maintaining high-resolution representations, while receiving rich contextual information from low-resolutions. 
The MRB consists of multiple (three in this paper) fully-convolutional streams connected in parallel. 
It allows information exchange across parallel streams in order to consolidate the high-resolution features with the help of low-resolution features, and vice versa. Next, we describe the individual components of MRB.
%

\vspace{0.4em} \noindent \textbf{Selective kernel feature fusion (SKFF).}
One fundamental property of neurons present in the visual cortex is to be able to change their receptive fields according to the stimulus \cite{li2019selective}. 
This mechanism of adaptively adjusting receptive fields can be incorporated in CNNs by using multi-scale feature generation (in the same layer) followed by feature aggregation and selection. 
The most commonly used approaches for feature aggregation include simple concatenation or summation. However, these choices provide limited expressive power to the network, as reported in \cite{li2019selective}.
In MRB, we introduce a nonlinear procedure for fusing features coming from multiple resolutions using a self-attention mechanism. Motivated by \cite{li2019selective}, we call it selective kernel feature fusion (SKFF). 

\begin{figure}[t]
\begin{center}
\scalebox{0.90}{
\begin{tabular}[t]{c} \hspace{-2mm}
\includegraphics[width=\textwidth]{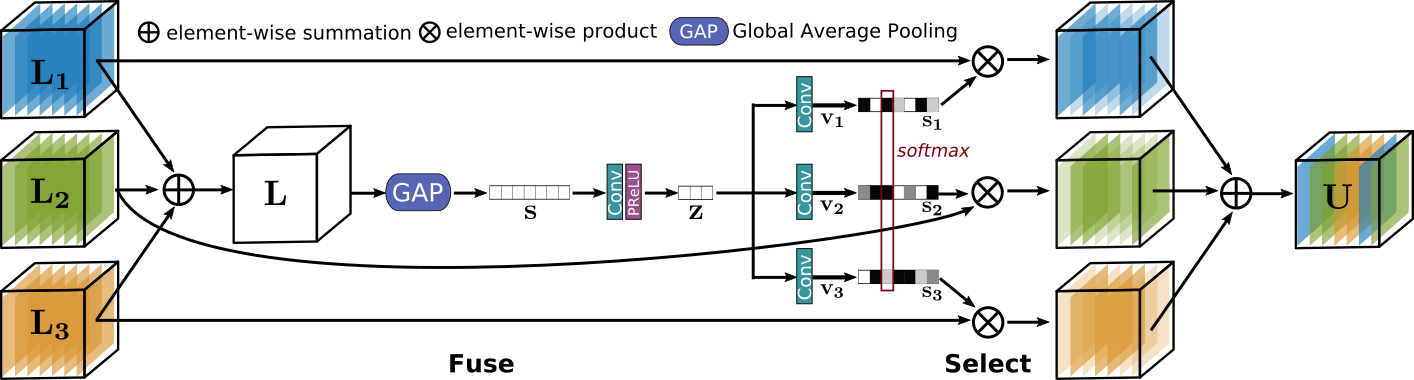}
\end{tabular}}
\end{center}
\vspace*{-6mm}
\caption{\small Schematic for selective kernel feature fusion (SKFF). It operates on features from multiple convolutional streams, and performs aggregation based on self-attention.}
\label{fig:skff}
\vspace*{-2em}
\end{figure}

The SKFF module performs dynamic adjustment of receptive fields via two operations --{\emph{Fuse} and \emph{Select}, as illustrated in Fig.~\ref{fig:skff}}.
The \emph{fuse} operator generates global feature descriptors by combining the information from multi-resolution streams. The \emph{select} operator uses these descriptors to recalibrate the feature maps (of different streams) followed by their aggregation. 
Next, we provide details of both operators for the three-stream case, but one can easily extend it to more streams. 
\textbf{(1) Fuse:} SKFF receives inputs from three parallel convolution streams carrying different scales of information. 
We first combine these multi-scale features using an element-wise sum as: $\mathbf{L = L_1 + L_2 + L_3}$. 
We then apply global average pooling (GAP) across the spatial dimension of $\mathbf{L} \in \mathbb{R}^{H\times W \times C}$ to compute channel-wise statistics $\mathbf{s} \in \mathbb{R}^{1\times 1 \times C}$. 
Next, we apply a channel-downscaling convolution layer to generate a compact feature representation $\mathbf{z} \in \mathbb{R}^{1\times 1 \times r}$, where $r=\frac{C}{8}$ for all our experiments.
Finally, the feature vector $\mathbf{z}$ passes through three parallel channel-upscaling convolution layers (one for each resolution stream) and provides us with three feature descriptors $\mathbf{v_1}, \mathbf{v_2}$ and $\mathbf{v_3}$, each with dimensions $1\times1\times C$. 
\textbf{(2) Select:} this operator applies the softmax function to $\mathbf{v_1}, \mathbf{v_2}$ and $\mathbf{v_3}$, yielding attention activations $\mathbf{s_1}, \mathbf{s_2}$ and $\mathbf{s_3}$ that we use to adaptively recalibrate multi-scale feature maps $\mathbf{L_1}, \mathbf{L_2}$ and $\mathbf{L_3}$, respectively. The overall process of feature recalibration and aggregation is defined as: $\mathbf{U = s_1 \cdot L_1 + s_2\cdot L_2 + s_3 \cdot L_3}$. 
Note that the SKFF uses $\sim6\times$ fewer parameters than aggregation with concatenation but generates more favorable results (an ablation study is provided in the experiments section).

\begin{figure}[t]
\begin{center}
\scalebox{0.99}{
\begin{tabular}[t]{c} \hspace{-2mm}
\includegraphics[width=\textwidth]{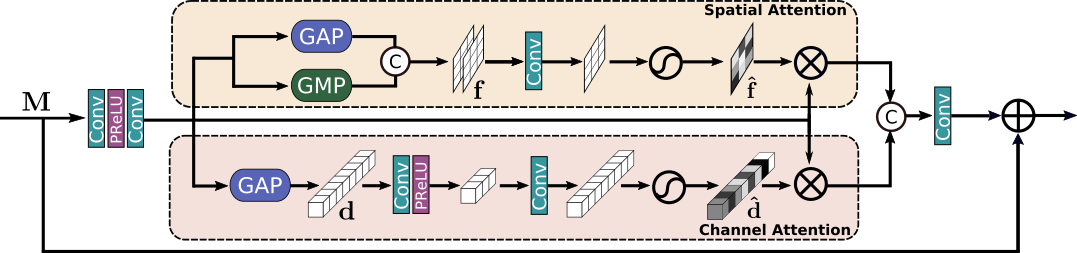}
\end{tabular}}
\end{center}
\vspace*{-6mm}
\caption{\small Dual attention unit incorporating spatial and channel attention mechanisms. }
\label{fig:dau}
\vspace{-0.6em}
\end{figure}

\vspace{0.4em} \noindent \textbf{Dual attention unit (DAU).}
While the SKFF block fuses information across multi-resolution branches, we also need a mechanism to share information within a feature tensor, both along the spatial and the channel dimensions.
Motivated by the advances of recent low-level vision methods~\cite{RCAN,RIDNet,dai2019second,zhang2019residual} based on the attention mechanisms~\cite{hu2018squeeze,wang2018non},
we propose the dual attention unit (DAU) to extract features in the convolutional streams. The schematic of DAU is shown in Fig.~\ref{fig:dau}.
The DAU suppresses less useful features and only allows more informative ones to pass further. 
This feature recalibration is achieved by using channel attention~\cite{hu2018squeeze} and spatial attention~\cite{woo2018cbam} mechanisms. 
\textbf{(1) Channel attention (CA)} branch exploits the inter-channel relationships of the convolutional feature maps by applying \emph{squeeze} and \emph{excitation} operations \cite{hu2018squeeze}. Given a feature map $\mathbf{M} \in \mathbb{R}^{H\times W \times C}$, the squeeze operation applies global average pooling across spatial dimensions to encode global context, thus yielding a feature descriptor $\mathbf{d} \in \mathbb{R}^{1\times 1 \times C}$. 
The excitation operator passes $\mathbf{d}$ through two convolutional layers followed by the sigmoid gating and generates activations $\mathbf{\hat{d}} \in \mathbb{R}^{1\times 1 \times C}$. 
Finally, the output of CA branch is obtained by rescaling $\mathbf{M}$ with the activations $\mathbf{\hat{d}}$.
\textbf{(2) Spatial attention (SA)} branch is designed to exploit the inter-spatial dependencies of convolutional features. The goal of SA is to generate a spatial attention map and use it to recalibrate the incoming features $\mathbf{M}$. 
To generate the spatial attention map, the SA branch first independently applies global average pooling and max pooling operations on features $\mathbf{M}$ along the channel dimensions and concatenates the outputs to form a feature map $\mathbf{f} \in \mathbb{R}^{H\times W \times 2}$. The map $\mathbf{f}$ is passed through a convolution and sigmoid activation to obtain the spatial attention map $\mathbf{\hat{f}} \in \mathbb{R}^{H\times W \times 1}$, which we then use to rescale $\mathbf{M}$. 

\begin{figure*}[t]
\centering
    \begin{subfigure}[t]{0.49\textwidth}
      \includegraphics[width=\textwidth]{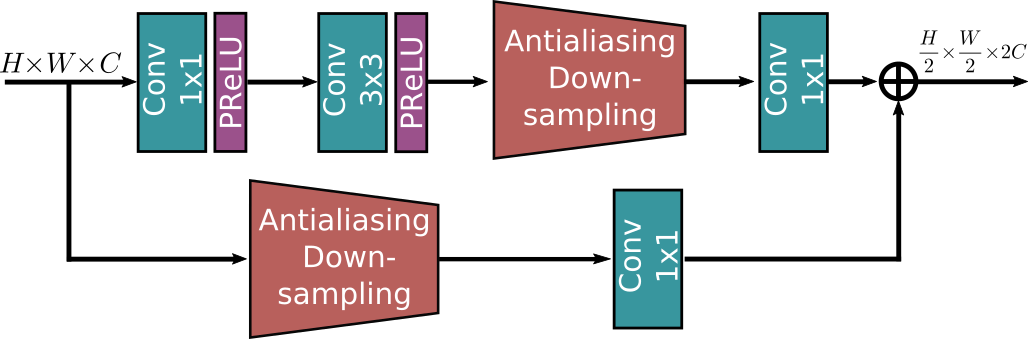}
      \caption{\small Downsampling module}
      \label{fig:downsample}
    \end{subfigure}
    \begin{subfigure}[t]{0.49\textwidth}
      \includegraphics[width=\textwidth]{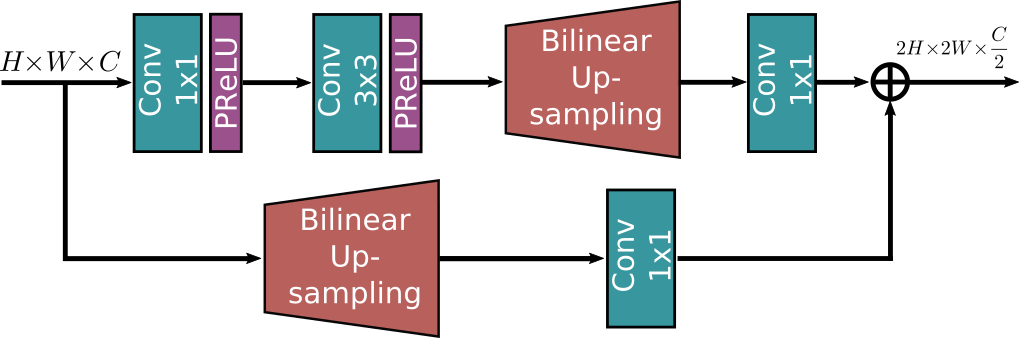}
      \caption{\small Upsampling module}
      \label{fig:upsample}
    \end{subfigure}
\vspace{-3mm}
\caption{\small Residual resizing modules to perform downsampling and upsampling. }
\vspace{-1.5em}
\end{figure*}

\vspace{0.2em} \noindent \textbf{Residual resizing modules.}
The proposed framework employs a recursive residual design (with skip connections) to ease the flow of information during the learning process. In order to maintain the residual nature of our architecture, we introduce residual resizing modules to perform downsampling (Fig.~\ref{fig:downsample}) and upsampling (Fig.~\ref{fig:upsample}) operations.   
In MRB, the size of feature maps remains constant along convolution streams.
On the other hand, across streams the feature map size changes depending on the input resolution index $i$ and the output resolution index $j$.
If $i<j$, the input feature tensor is downsampled, and if $i>j$, the feature map is upsampled.
To perform $2\times$ downsampling (halving the spatial dimension and doubling the channel dimension), we apply the module in Fig.~\ref{fig:downsample} only once. For $4\times$ downsampling, the module is applied twice, consecutively.
Similarly, one can perform $2\times$ and $4\times$ upsampling by applying the module in Fig.~\ref{fig:upsample} once and twice, respectively. 
Note in Fig.~\ref{fig:downsample}, we integrate anti-aliasing downsampling~\cite{zhang2019making} to improve the shift-equivariance of our network.

%

\section{Experiments}
In this section, we perform qualitative and quantitative assessment of the results produced by our MIRNet and compare it with the previous best methods. 
Next, we describe the datasets, and then provide the implementation details. Finally, we report results for \textbf{(a)} image denoising, \textbf{(b)} super-resolution and \textbf{(c)} image enhancement on five real image datasets. 
%

\subsection{Real Image Datasets}
\noindent \textbf{Image denoising.} 
\noindent \textbf{(1) DND \cite{dnd}} consists of $50$ images captured with four consumer cameras.  
Since the images are of very high-resolution, the dataset providers extract $20$ crops of size $512\times512$ from each image, yielding $1000$ patches in total. All these patches are used for testing (as DND does not contain training or validation sets).  
The ground-truth noise-free images are not released publicly, therefore the image quality scores in terms of PSNR and SSIM can only be obtained through an online server \cite{dndwebsite}. 
\noindent \textbf{(2) SIDD \cite{sidd}} is particularly collected with smartphone cameras. Due to the small sensor and high-resolution, the noise levels in smartphone images are much higher than those of DSLRs. 
SIDD contains $320$ image pairs for training and $1280$ for validation. 

\vspace{0.4em}\noindent \textbf{Super-resolution.} 
\noindent \textbf{(1) RealSR~\cite{RealSR}} contains real-world LR-HR image pairs of the same scene captured by adjusting the focal-length of the cameras. 
RealSR has both indoor and outdoor images taken with two cameras. 
The number of training image pairs for scale factors $\times2$, $\times3$ and $\times4$ are $183$, $234$ and $178$, respectively. For each scale factor, $30$ test images are also provided in RealSR.


\vspace{0.4em}\noindent \textbf{Image enhancement.} 
\noindent \textbf{(1)~LoL~\cite{wei2018deep}} is created for low-light image enhancement problem. It provides 485 images for training and 15 for testing. Each image pair in LoL consists of a low-light input image and its corresponding well-exposed reference image.
\noindent \textbf{(2)~MIT-Adobe FiveK \cite{mit_fivek}} contains $5000$ images of various indoor and outdoor scenes captured with DSLR cameras in different lighting conditions. 
The tonal attributes of all images are manually adjusted by five different trained photographers (labelled as experts A to E). 
Same as in \cite{hu2018exposure,park2018distort,wang2019underexposed}, we also consider the enhanced images of expert C as the ground-truth. 
Moreover, the first 4500 images are used for training and the last 500 for testing.

\subsection{Implementation Details}
The proposed architecture is end-to-end trainable and requires no pre-training of sub-modules. 
We train three different networks for three different restoration tasks. The training parameters, common to all experiments, are the following. 
We use 3 RRGs, each of which further contains $2$ MRBs. The MRB consists of $3$ parallel streams with channel dimensions of $64, 128, 256$ at resolutions $1, \frac{1}{2}, \frac{1}{4}$, respectively. Each stream has $2$ DAUs.  
The models are trained with the Adam optimizer ($\beta_1 = 0.9$, and $\beta_2=0.999$) for $7\times10^5$ iterations. The initial learning rate is set to $2\times10^{-4}$. We employ the cosine annealing strategy~\cite{loshchilov2016sgdr} to steadily decrease the learning rate from initial value to $10^{-6}$ during training. 
We extract patches of size $128\times128$ from training images. The batch size is set to $16$ and, for data augmentation, we perform horizontal and vertical flips.


\begin{table*}[t]
\begin{center}
\caption{\small Denoising comparisons on the SIDD dataset \cite{sidd}.}
\label{table:sidd}
\setlength{\tabcolsep}{2.5pt}
\scalebox{0.70}{
\begin{tabular}{l c c c c c c c c c c c c c c c}
\toprule
 \rowcolor{color3} Method & DnCNN & MLP   & GLIDE & TNRD   & FoE    & BM3D    & WNNM    & NLM    & KSVD    & EPLL    & CBDNet & RIDNet &  VDN &
 MIRNet \\
\rowcolor{color3} & \cite{DnCNN} & \cite{MLP} & \cite{GLIDE} & \cite{TNRD} & \cite{FoE} & \cite{BM3D} &  \cite{WNNM}	 & \cite{NLM} &  \cite{KSVD} & \cite{EPLL} & \cite{CBDNet}& \cite{RIDNet} & \cite{VDN} & (Ours)
\\
\midrule
PSNR~$\textcolor{black}{\uparrow}$ &  23.66  &   24.71  &    24.71  &  24.73  &  25.58  & 25.65  &   25.78  &   26.76  &  26.88  & 27.11  &  30.78  & 38.71 & 39.28 &
\textbf{39.72}\\
SSIM~$\textcolor{black}{\uparrow}$ &  0.583 &  0.641 &  0.774 &  0.643 &  0.792 &  0.685 &  0.809 &  0.699 &  0.842 &  0.870 & 0.754   &    0.914 &  0.909  & 
\textbf{0.959}\\
\bottomrule
\end{tabular}}
\end{center}\vspace{-2.5em}
\end{table*}

\begin{table*}[t]
\begin{center}
\caption{\small Denoising comparisons on the DND dataset \cite{dnd}.}
\label{table:dnd}
\setlength{\tabcolsep}{1.4pt}
\scalebox{0.70}{
\begin{tabular}{l c c c c c c c c c c c c c c c}
\toprule
 \rowcolor{color3} Method & EPLL   & TNRD & MLP   & BM3D & FoE 	& WNNM & KSVD  	& MCWNNM & FFDNet+ & TWSC  & CBDNet & RIDNet  & VDN &
 MIRNet \\
\rowcolor{color3} & \cite{EPLL} & \cite{TNRD} &  \cite{MLP} & \cite{BM3D} & \cite{FoE} &  \cite{WNNM}	 & \cite{KSVD} &  \cite{MCWNNM} & \cite{FFDNetPlus} & \cite{TWSC}& \cite{CBDNet} & \cite{RIDNet} & \cite{VDN}
& (Ours)\\
\midrule
PSNR~$\textcolor{black}{\uparrow}$ & 33.51 &  33.65  & 34.23 & 34.51 & 34.62 &  34.67 & 36.49 &  37.38 & 37.61 & 37.94 & 38.06 &  39.26 & 39.38 & 
\textbf{39.88}\\
SSIM~$\textcolor{black}{\uparrow}$& 0.824 & 0.831 & 0.833 & 0.851 & 0.885 & 0.865 & 0.898 & 0.929 & 0.942 & 0.940 & 0.942 & 0.953 & 0.952 & 
\textbf{0.956}\\
\bottomrule
\end{tabular}}
\end{center}\vspace{-1em}
\end{table*}

\begin{figure}[!t]
\begin{center}
\tiny
\scalebox{1}{
\begin{tabular}[b]{c@{ } c@{ }  c@{ } c@{ } c@{ }	}\hspace{-2.4mm}
    \multirow{4}{*}{\includegraphics[width=.35\textwidth,valign=t]{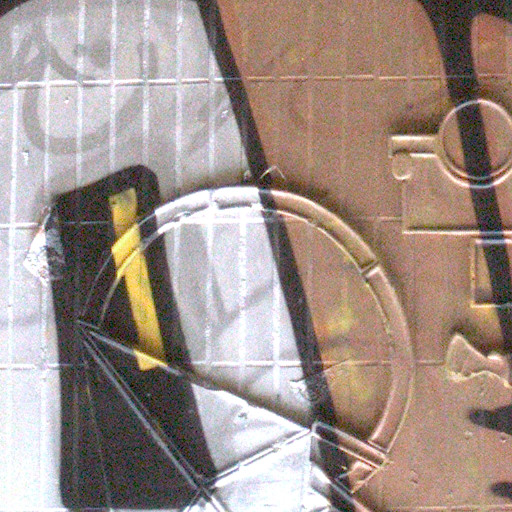}} 
    &  \includegraphics[trim={10.8cm 6.5cm  3cm  7.2cm },clip,width=.152\textwidth,valign=t]{Images/Denoising/DND/RGB/noisy_26_90}
    & \includegraphics[trim={10.8cm 6.5cm  3cm  7.2cm },clip,width=.152\textwidth,valign=t]{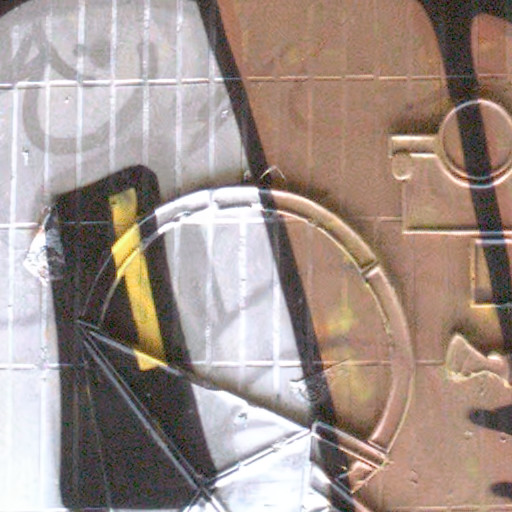}
    & \includegraphics[trim={10.8cm 6.5cm  3cm  7.2cm },clip,width=.152\textwidth,valign=t]{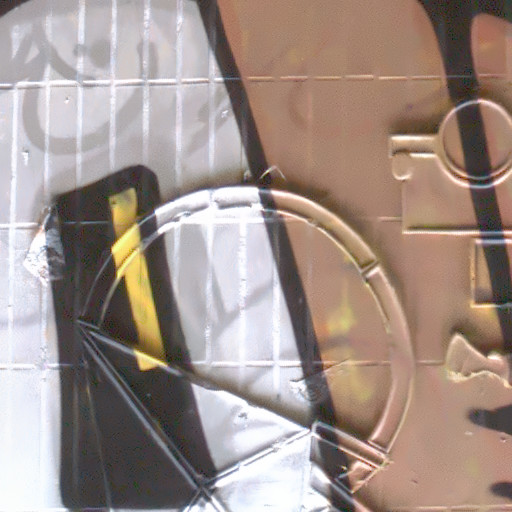}
     & \includegraphics[trim={10.8cm 6.5cm  3cm  7.2cm },clip,width=.152\textwidth,valign=t]{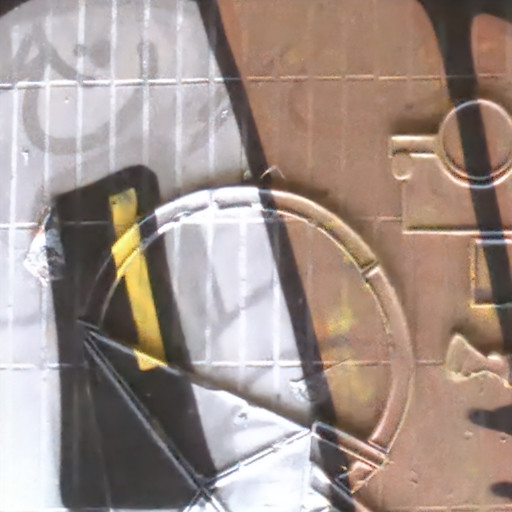}
    \vspace{0.5mm}
\\
    &  26.90 dB     &30.91 dB  & 32.47 dB
    & 33.29 \\
    & Noisy &BM3D~\cite{BM3D}  & NC~\cite{lebrun2015NC} & DnCNN~\cite{DnCNN}
    \\
    &\includegraphics[trim={10.8cm 6.5cm  3cm  7.2cm },clip,width=.152\textwidth,valign=t]{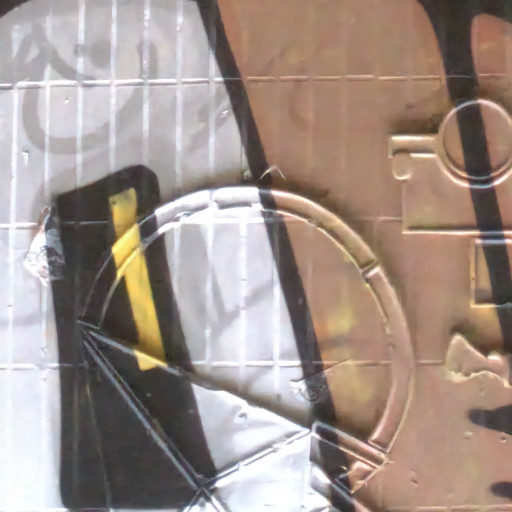}  
    &\includegraphics[trim={10.8cm 6.5cm  3cm  7.2cm },clip,width=.152\textwidth,valign=t]{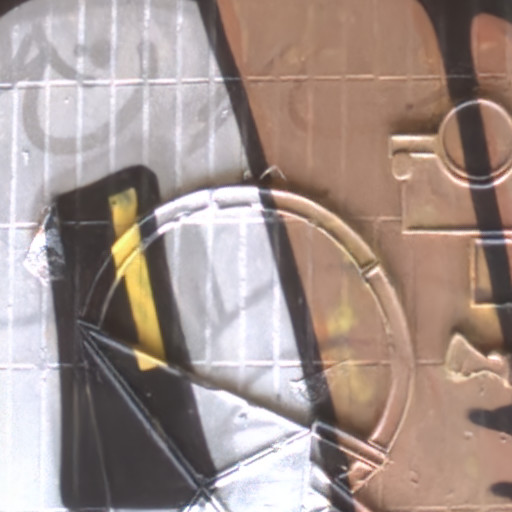}&
     \includegraphics[trim={10.8cm 6.5cm  3cm  7.2cm },clip,width=.152\textwidth,valign=t]{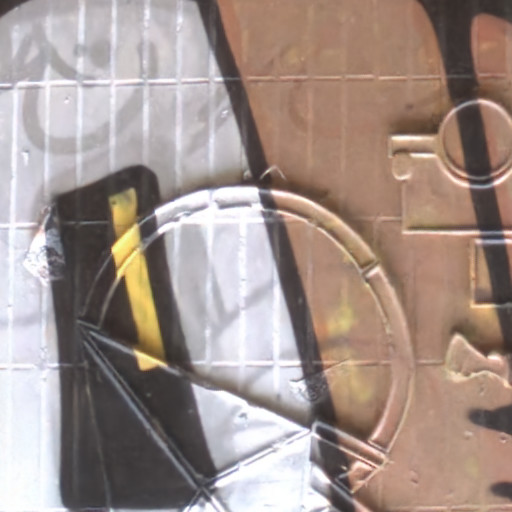}&
     \includegraphics[trim={10.8cm 6.5cm  3cm  7.2cm },clip,width=.152\textwidth,valign=t]{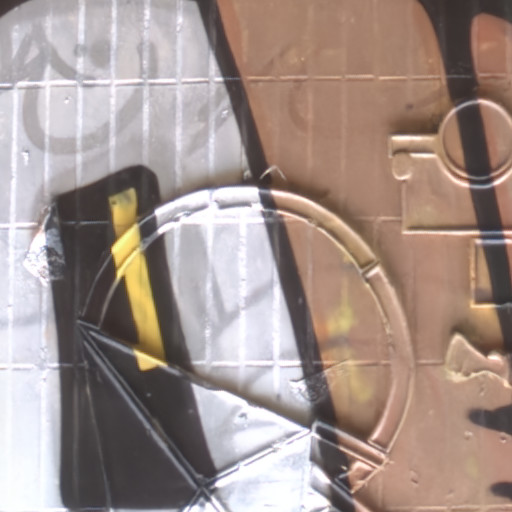}
     \vspace{0.5mm}
     \\

     26.90 dB & 33.62 dB 
     & 33.89 dB
     & 34.09 dB & \textbf{34.77 dB}\\
           Noisy Image  & CBDNet~\cite{CBDNet} & VDN~\cite{VDN}      
           & RIDNet~\cite{RIDNet}     & MIRNet (Ours) \\
\end{tabular}}
\end{center}
\vspace{-6mm}
\caption{\small Denoising example from DND~\cite{dnd}. Our MIRNet generates visually-pleasing and artifact-free results. }
\label{fig:dnd example}
\vspace{-1.5em}
\end{figure}

\vspace{-0.4em}
\subsection{Image Denoising}
In this section, we demonstrate the effectiveness of the proposed MIRNet for image denoising.
We train our network only on the training set of the SIDD~\cite{sidd} and directly evaluate it on the test images of both SIDD and DND~\cite{dnd} datasets.
Quantitative comparisons in terms of PSNR and SSIM metrics are summarized in Table~\ref{table:sidd} and Table~\ref{table:dnd} for SIDD and DND, respectively.
Both tables show that our MIRNet performs favourably against the data-driven, as well as conventional, denoising algorithms. 
Specifically, when compared to the recent best method VDN~\cite{VDN}, our algorithm demonstrates a performance gain of $0.44$~dB on SIDD and $0.50$~dB on DND.  
Furthermore, it is worth noting that CBDNet~\cite{CBDNet} and RIDNet~\cite{RIDNet} use additional training data, yet our method provides significantly better results. For instance, our method achieves $8.94$~dB improvement over CBDNet~\cite{CBDNet} on the SIDD dataset and $1.82$~dB on DND.

\begin{figure}[t]
\begin{center}
\tiny
\begin{tabular}[t]{c@{ }c@{ }c@{ }c@{ }c@{ }c@{ }c@{ }}
\includegraphics[width=.16\textwidth]{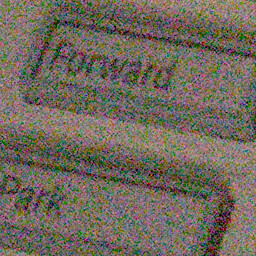}& \hspace{-1.4mm}
\includegraphics[width=.16\textwidth]{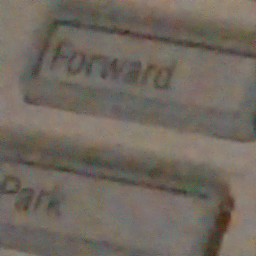}&  \hspace{-1.4mm}
\includegraphics[width=.16\textwidth]{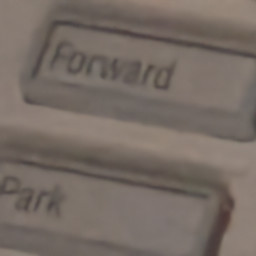}& \hspace{-1.4mm}
\includegraphics[width=.16\textwidth]{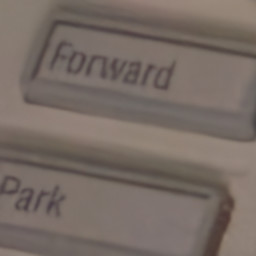}& \hspace{-1.4mm}
\includegraphics[width=.16\textwidth]{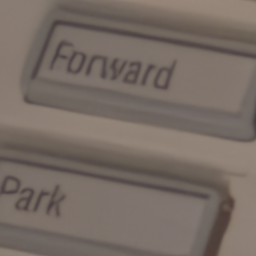}&  \hspace{-1.4mm}
\includegraphics[width=.16\textwidth]{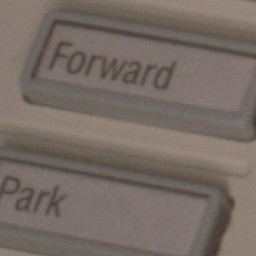}
\vspace{0.2mm}
\\
\vspace{0.5mm}
 18.25 dB & 28.84 dB & 35.57 dB & 36.39 dB & \textbf{36.97 dB} &\\
\includegraphics[width=.16\textwidth]{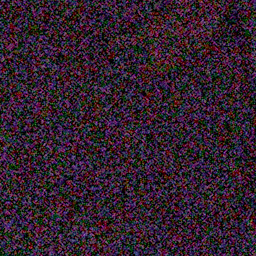}& \hspace{-1.4mm}
\includegraphics[width=.16\textwidth]{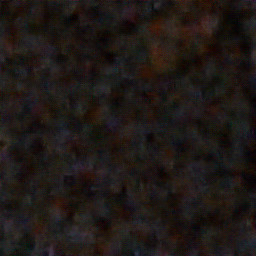}&  \hspace{-1.4mm}
\includegraphics[width=.16\textwidth]{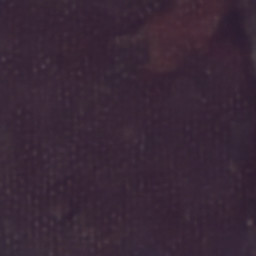}& \hspace{-1.4mm}
\includegraphics[width=.16\textwidth]{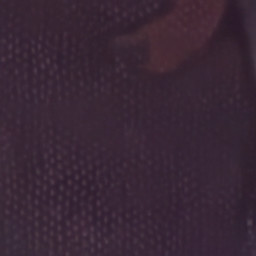}& \hspace{-1.4mm}
\includegraphics[width=.16\textwidth]{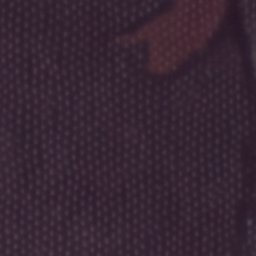}&  \hspace{-1.4mm}
 \includegraphics[width=.16\textwidth]{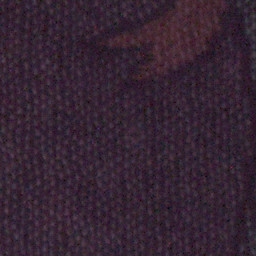}
\vspace{0.2mm}
\\
\vspace{0.5mm}
 18.16 dB & 20.36 dB & 29.83 dB & 30.31 dB & \textbf{31.36 dB} &\\
 Noisy
& CBDNet~\cite{CBDNet}  & RIDNet~\cite{RIDNet} & VDN~\cite{VDN} & MIRNet (Ours) & Reference \\
\end{tabular}
\end{center}
\vspace{-6mm}
\caption{\small Denoising examples from SIDD~\cite{sidd}. Our method effectively removes real noise from challenging images, while better recovering structural content and fine texture.}
\label{fig:sidd example}
\vspace{-1.2em}
\end{figure}

In Fig.~\ref{fig:dnd example} and Fig.~\ref{fig:sidd example}, we present visual comparisons of our results with those of other competing algorithms. 
It can be seen that our MIRNet is effective in removing real noise and produces perceptually-pleasing and sharp images.
Moreover, it is capable of maintaining the spatial smoothness of the homogeneous regions without introducing artifacts.
In contrast, most of the other methods either yield over-smooth images and thus sacrifice structural content and fine textural details, or produce images with chroma artifacts and blotchy texture.

\begin{table}[t]
\begin{center}
\caption{\small Super-resolution evaluation on the RealSR~\cite{RealSR} dataset. Compared to the state-of-the-art, our method consistently yields significantly better image quality scores for all three scaling factors.}
\label{table:realSR}
\setlength{\tabcolsep}{7.1pt}
\scalebox{0.70}{
\begin{tabular}{l | c c | c c | c c | c c| c c| c c}
\toprule
\rowcolor{color3} & \multicolumn{2}{c|}{Bicubic} & \multicolumn{2}{c|}{VDSR~\cite{VDSR}} & \multicolumn{2}{c|}{SRResNet~\cite{SRResNet}} & \multicolumn{2}{c|}{RCAN~\cite{RCAN}} & \multicolumn{2}{c|}{LP-KPN~\cite{RealSR}} & \multicolumn{2}{c}{MIRNet (Ours)} \\
\cline{2-13}
\rowcolor{color3} Scale & PSNR & SSIM & PSNR & SSIM & PSNR & SSIM & PSNR & SSIM & PSNR & SSIM & PSNR & SSIM \\
\midrule
$\times$2 &   32.61  &  0.907   &   33.64  &  0.917 &   33.69  &  0.919   &  33.87  & 0.922 &  33.90  &  0.927   &   \textbf{34.35}  & \textbf{0.935} \\
$\times$3 &   29.34 &  0.841   &   30.14  & 0.856 &   30.18  &  0.859   &  30.40  & 0.862 &  30.42  &  0.868   &   \textbf{31.16}  &  \textbf{0.885}\\
$\times$4 &   27.99  &  0.806   &  28.63  &  0.821 &   28.67  &  0.824   &  28.88  & 0.826 & 28.92  &  0.834   &   \textbf{29.14}  &  \textbf{0.843} \\
\bottomrule
\end{tabular}}
\end{center}
\vspace{-1.8em}
\end{table}

\vspace{0.4em}\noindent \textbf{Generalization capability.} The DND and SIDD datasets are acquired with different sets of cameras having different noise characteristics. Since the DND benchmark does not provide training data, setting a new state-of-the-art on DND with our SIDD trained network indicates the good generalization capability of our approach.


\subsection{Super-Resolution (SR)}

We compare our MIRNet against the state-of-the-art SR algorithms (VDSR~\cite{VDSR}, SRResNet~\cite{SRResNet}, RCAN~\cite{RCAN}, LP-KPN~\cite{RealSR}) on the testing images of the RealSR~\cite{RealSR} for upscaling factors of $\times2$, $\times3$ and $\times4$. 
Note that all the benchmarked algorithms are trained on the RealSR~\cite{RealSR} dataset for a fair comparison. 
In the experiments, we also include bicubic interpolation~\cite{keys1981cubic}, which is the most commonly used method for generating super-resolved images. 
Here, we compute the PSNR and SSIM scores using the Y channel (in YCbCr color space), as it is a common practice in the SR literature~\cite{RCAN,RealSR,wang2019deep,anwar2019deep}. 
The results in Table~\ref{table:realSR} show that the bicubic interpolation provides the least accurate results, thereby indicating its low suitability for dealing with real images. 
Moreover, the same table shows that the recent method LP-KPN~\cite{RealSR} provides marginal improvement of only $\sim0.04$~dB over the previous best method RCAN~\cite{RCAN}.
In contrast, our method significantly advances state-of-the-art and consistently yields better image quality scores than other approaches for all three scaling factors.
Particularly, compared to LP-KPN~\cite{RealSR}, our method provides performance gains of $0.45$~dB, $0.74$~dB, and $0.22$~dB for scaling factors $\times2$, $\times3$ and $\times4$, respectively. The trend is similar for the SSIM metric as well.   

\begin{figure}[t]
\begin{center}
\tiny
\scalebox{1}{
\begin{tabular}[b]{c@{ } c@{ }  c@{ } c@{ } c@{ }	}\hspace{-1.5mm}
    \multirow{3}{*}{\includegraphics[width=.333\textwidth,valign=t]{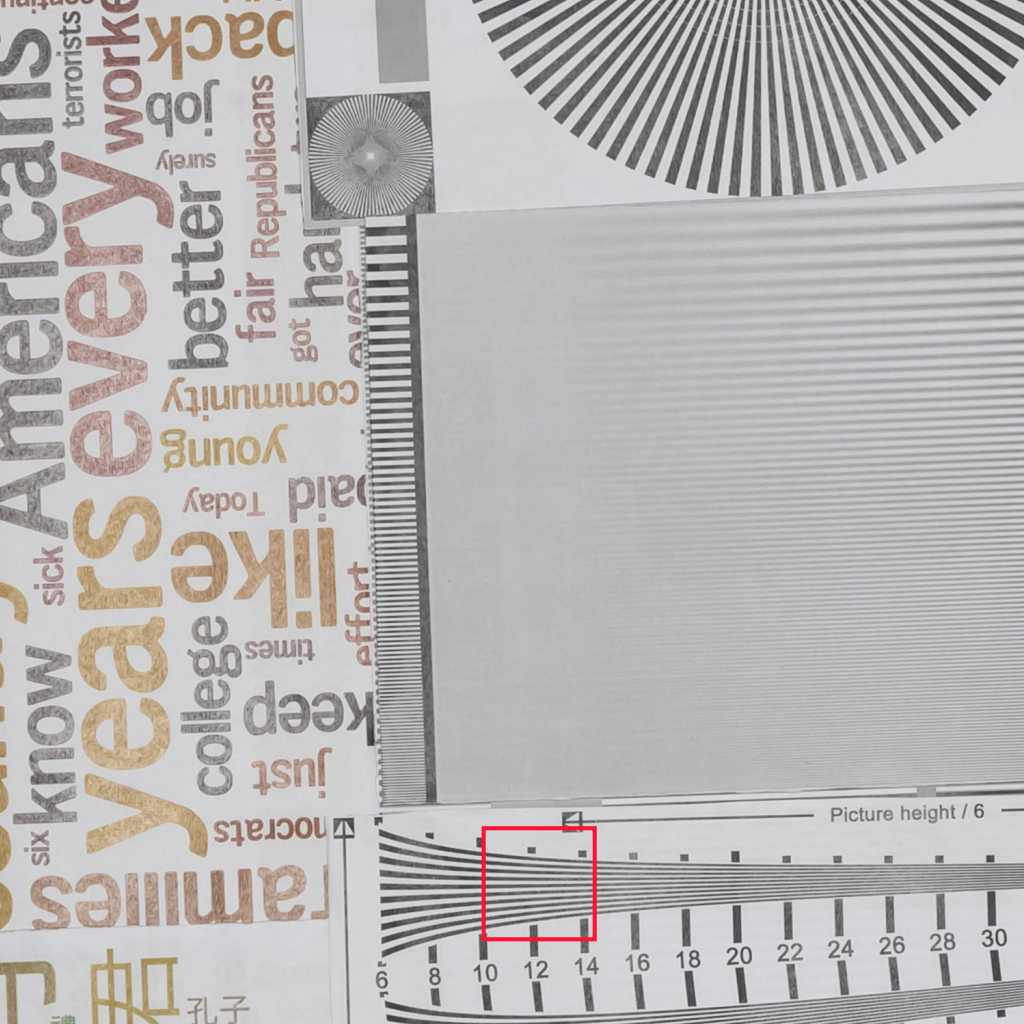}} 
    &  \includegraphics[width=.154\textwidth,valign=t]{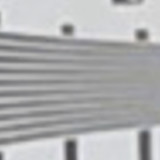}
    & \includegraphics[width=.154\textwidth,valign=t]{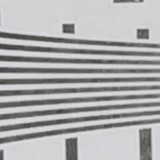}
    & \includegraphics[width=.154\textwidth,valign=t]{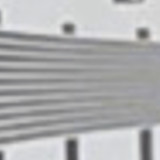}
    &\includegraphics[width=.154\textwidth,valign=t]{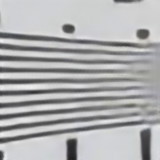}
    \vspace{0.5mm}
    \\
    \vspace{0.5mm}
    & LR & HR  & Bicubic & SRResNet~\cite{SRResNet} 
    \\
    &\includegraphics[width=.154\textwidth,valign=t]{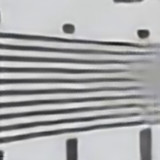}  
    &\includegraphics[width=.154\textwidth,valign=t]{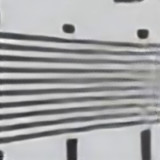}
    &\includegraphics[width=.154\textwidth,valign=t]{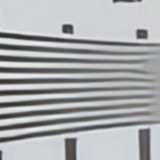}
    &\includegraphics[width=.154\textwidth,valign=t]{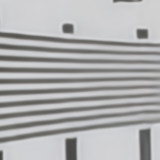}
    \vspace{0.5mm}
    \\
    Image  & VDSR~\cite{VDSR} & RCAN~\cite{RCAN} & LP-KPN~\cite{RealSR} & MIRNet (Ours) \\
\end{tabular}}
\end{center}
\vspace{-6mm}
\caption{\small Comparisons for $\times4$ super-resolution from the RealSR~\cite{RealSR} dataset. The image produced by our MIRNet is more faithful to the ground-truth than other competing methods (see lines near the right edge of the crops).}
\label{fig:sr example}
\vspace{-0.5em}
\end{figure}

\begin{figure}[t]
\begin{center}
\tiny
\scalebox{1}{
\begin{tabular}[b]{c@{ } c@{ }  c@{ } c@{ } c@{ } c@{ } }
    \hspace{-3mm}
    \multirow{5}{*}{HR}
    &  \includegraphics[trim={1100 550 150 900 },clip,width=.177\textwidth,valign=t]{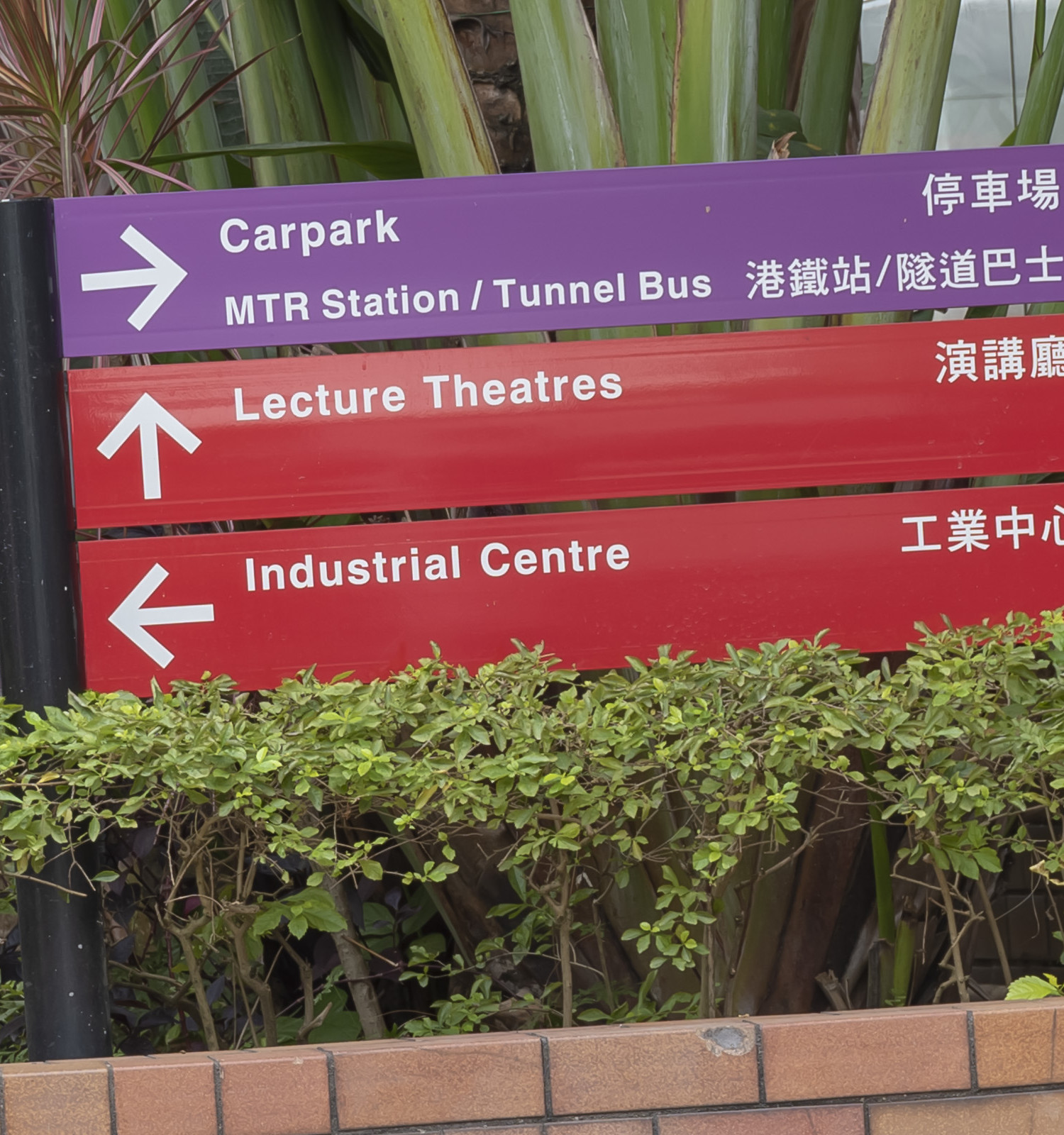}\vspace{1.2mm}
    &  \includegraphics[trim={320 180 930 870 },clip,width=.177\textwidth,valign=t]{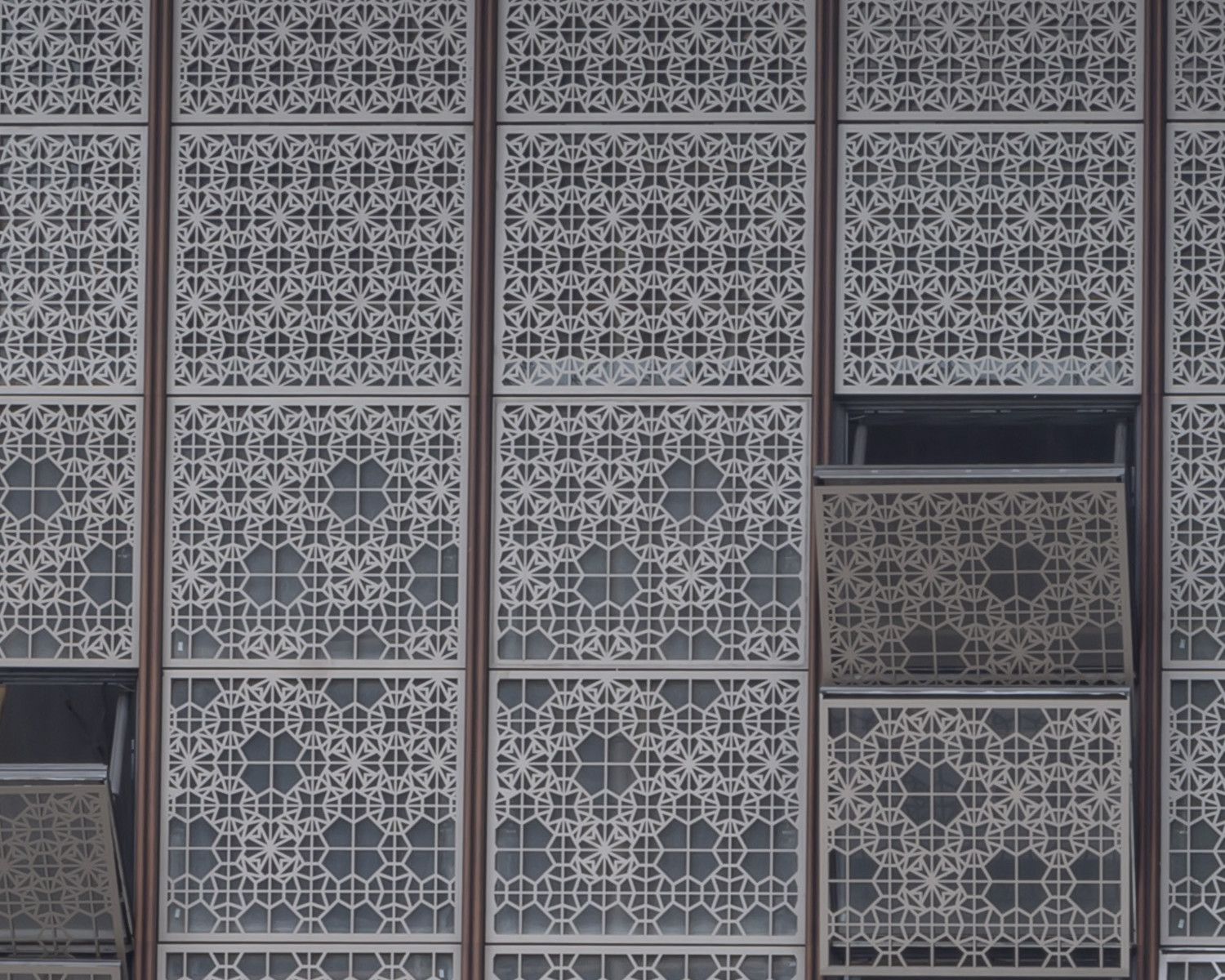}
    &  \includegraphics[trim={250 100 300 850 },clip,width=.177\textwidth,valign=t]{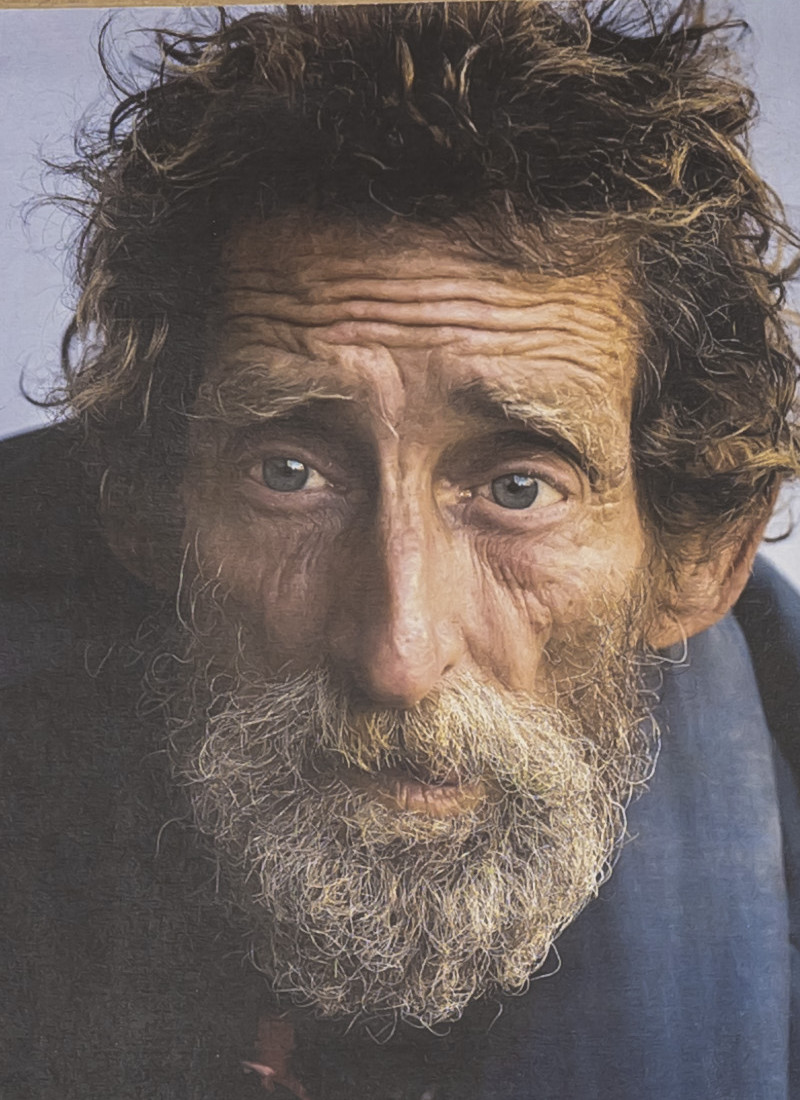}
    &  \includegraphics[trim={350 550 700 300 },clip,width=.177\textwidth,valign=t]{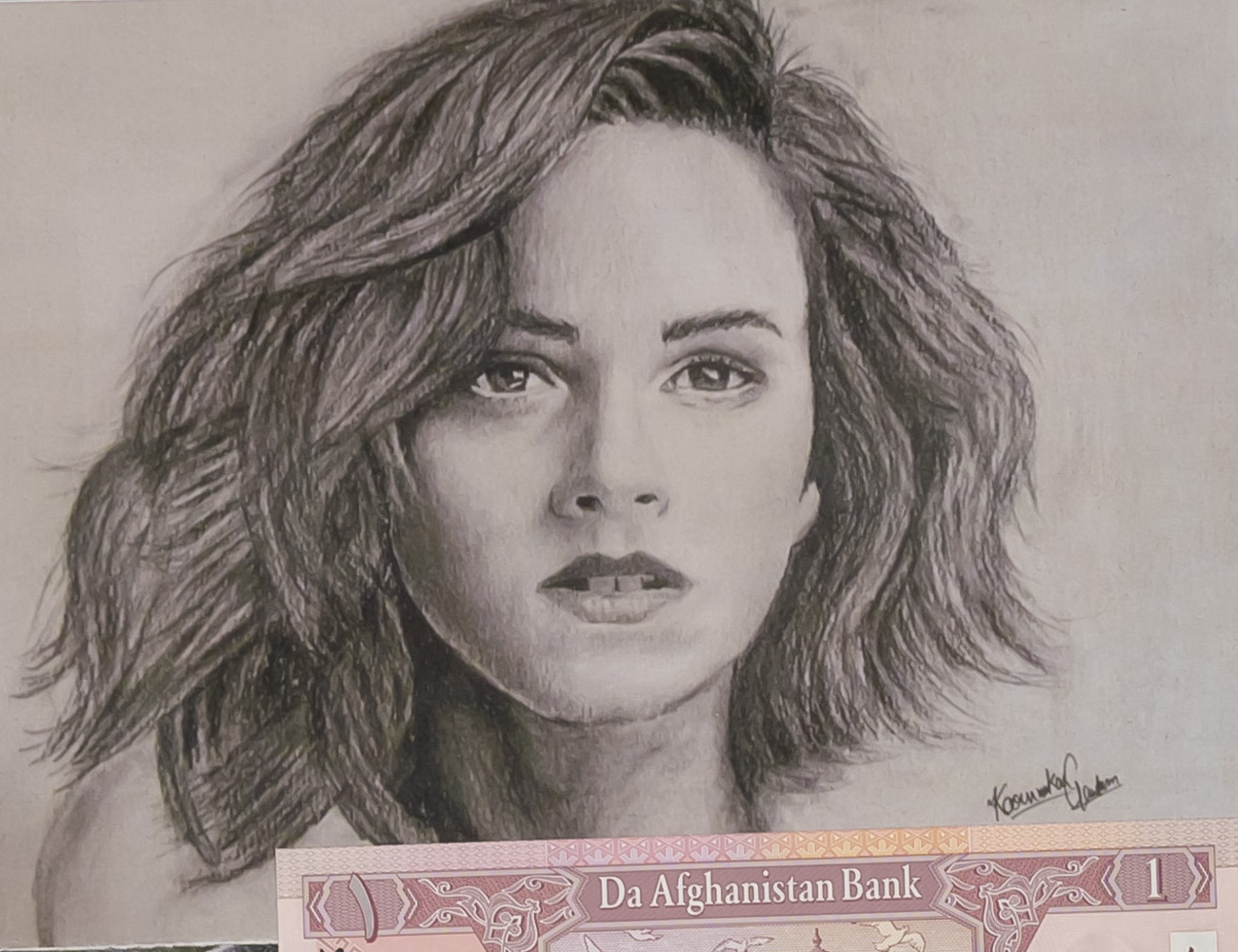}
    &  \includegraphics[trim={150 250 1700 1400 },clip,width=.177\textwidth,valign=t]{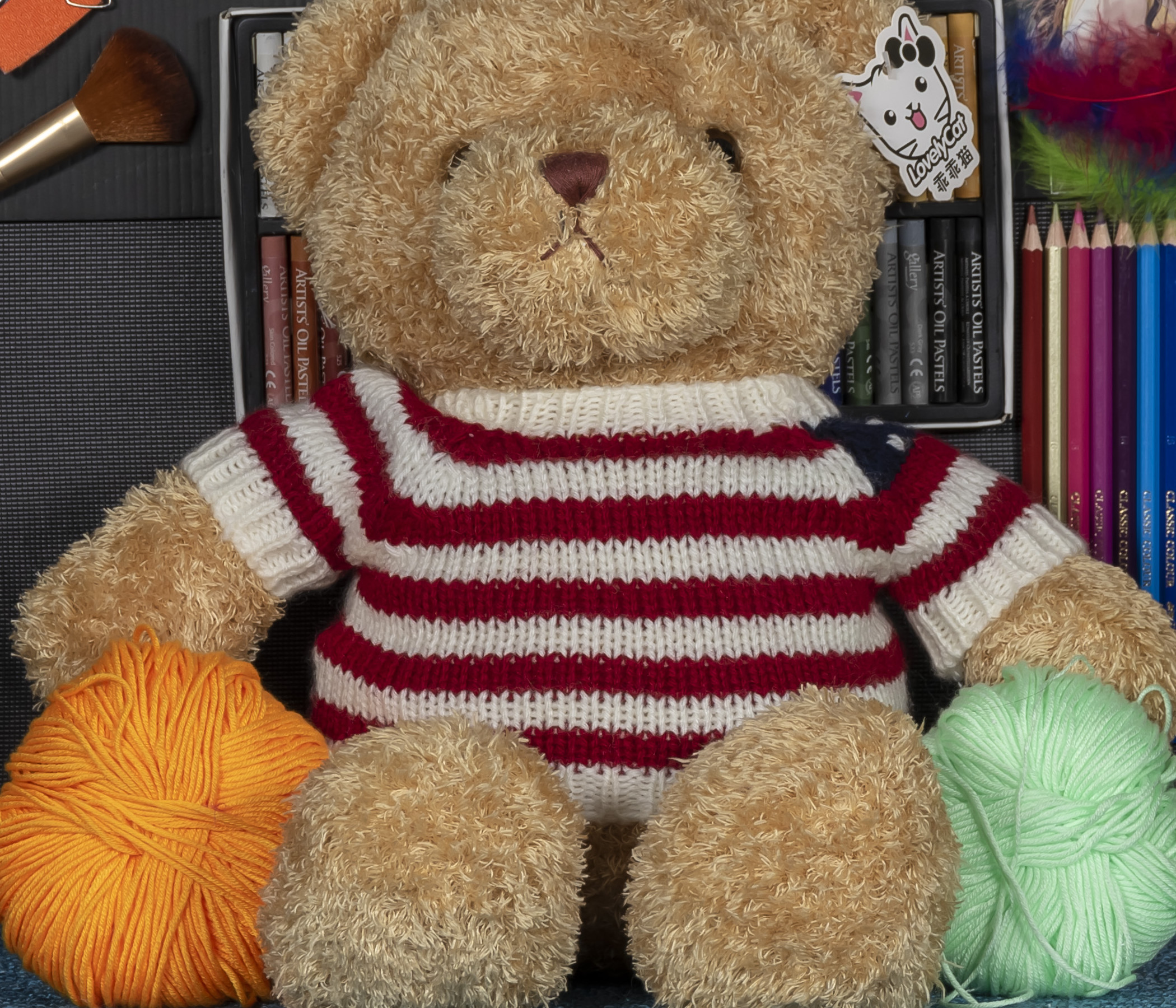}
    \\
    \hspace{-3mm}
    \multirow{6}{*}{\makecell{LP-KPN \\ \cite{RealSR}}}
    &  \includegraphics[trim={1100 550 150 900 },clip,width=.177\textwidth,valign=t]{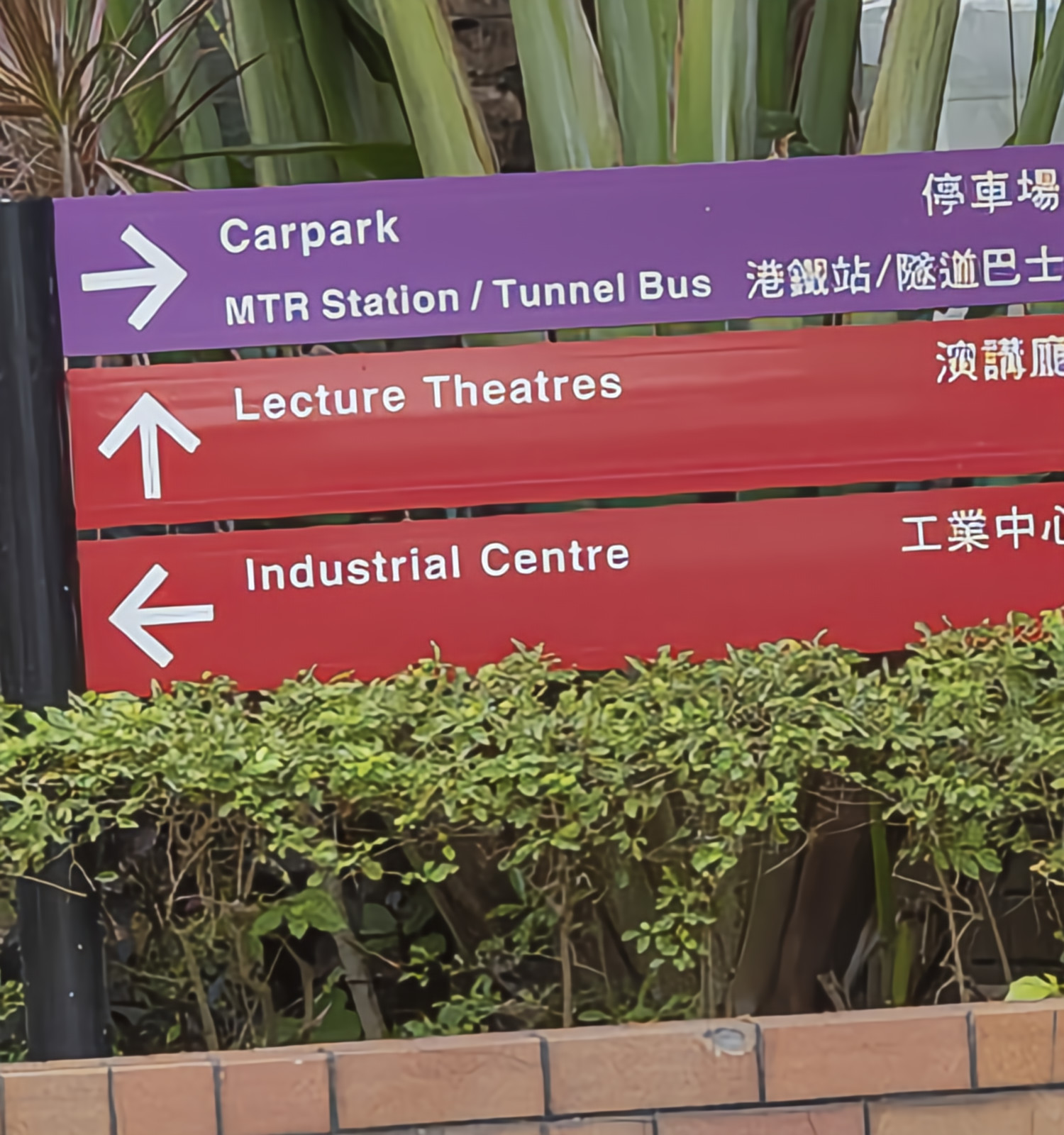}
    &  \includegraphics[trim={320 180 930 870 },clip,width=.177\textwidth,valign=t]{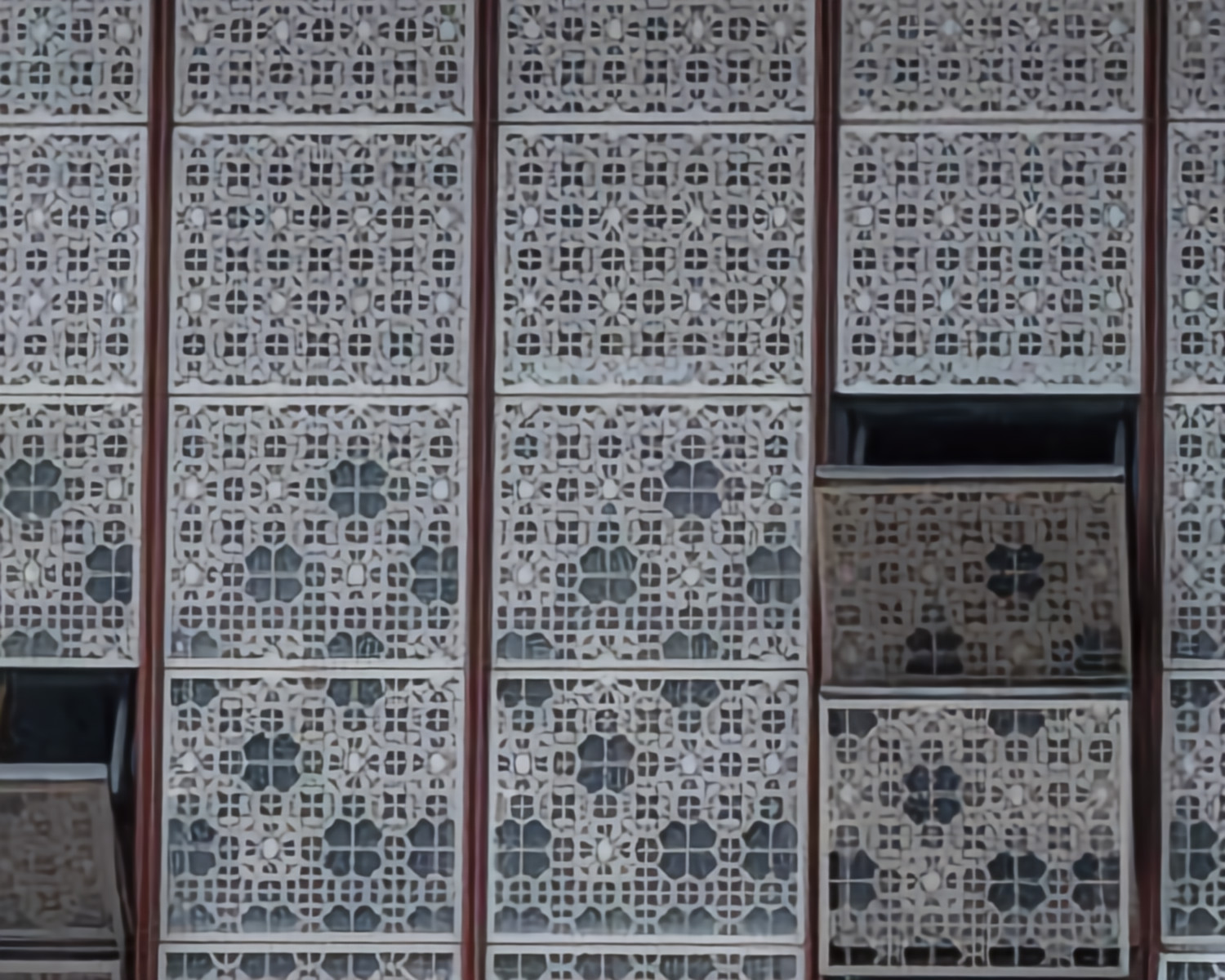}
    &  \includegraphics[trim={250 100 300 850 },clip,width=.177\textwidth,valign=t]{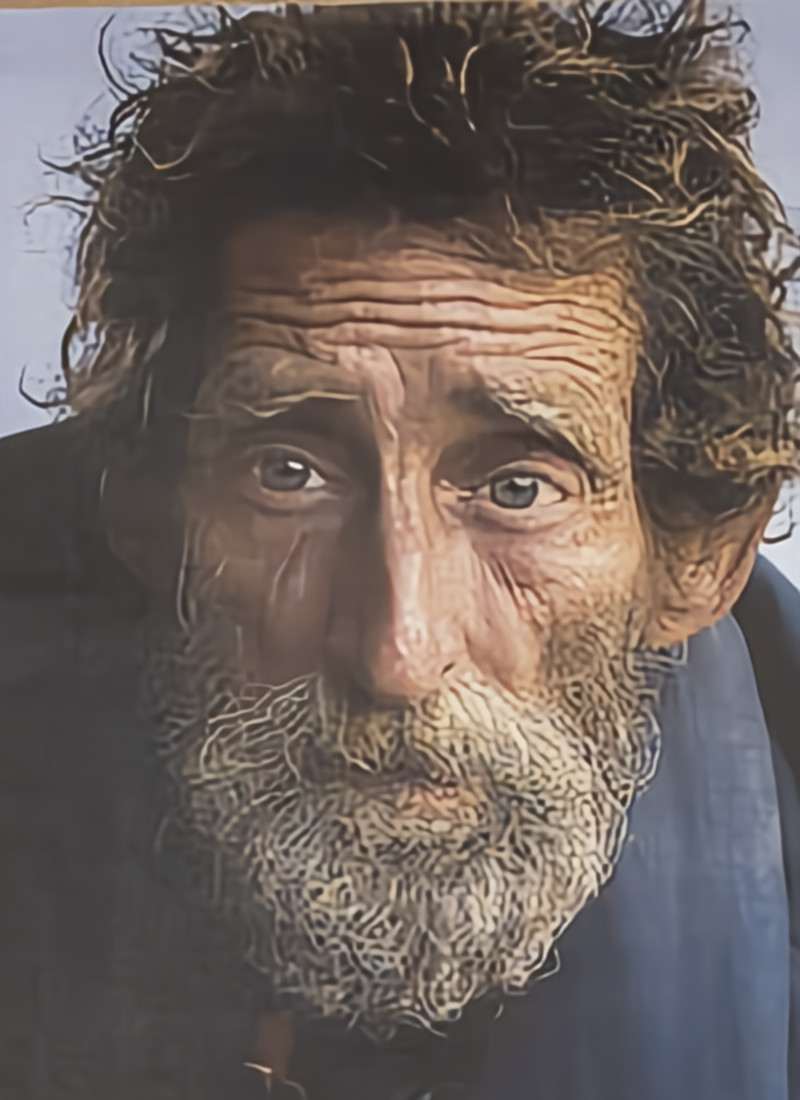}
    &  \includegraphics[trim={350 550 700 300 },clip,width=.177\textwidth,valign=t]{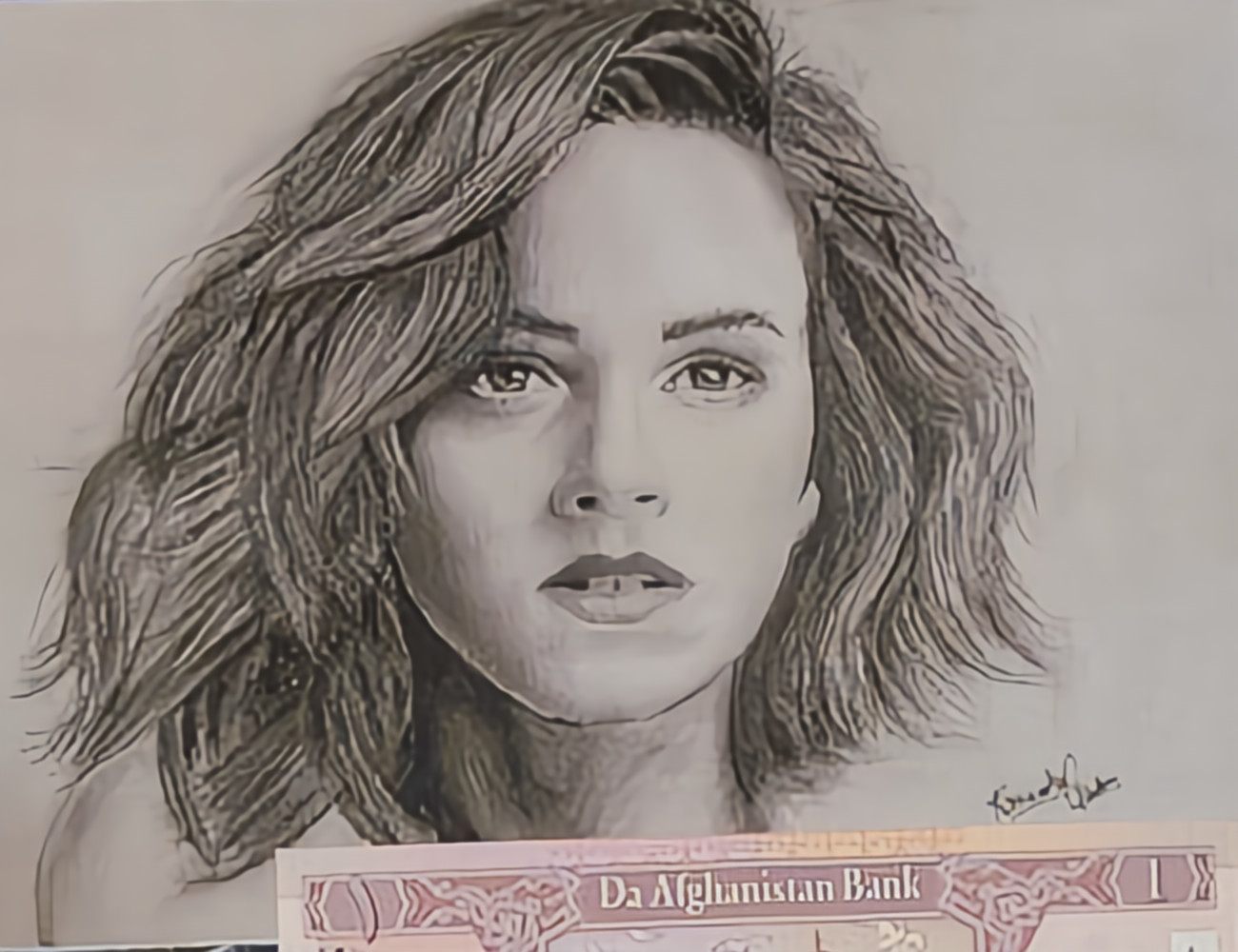}
    &  \includegraphics[trim={150 250 1700 1400 },clip,width=.177\textwidth,valign=t]{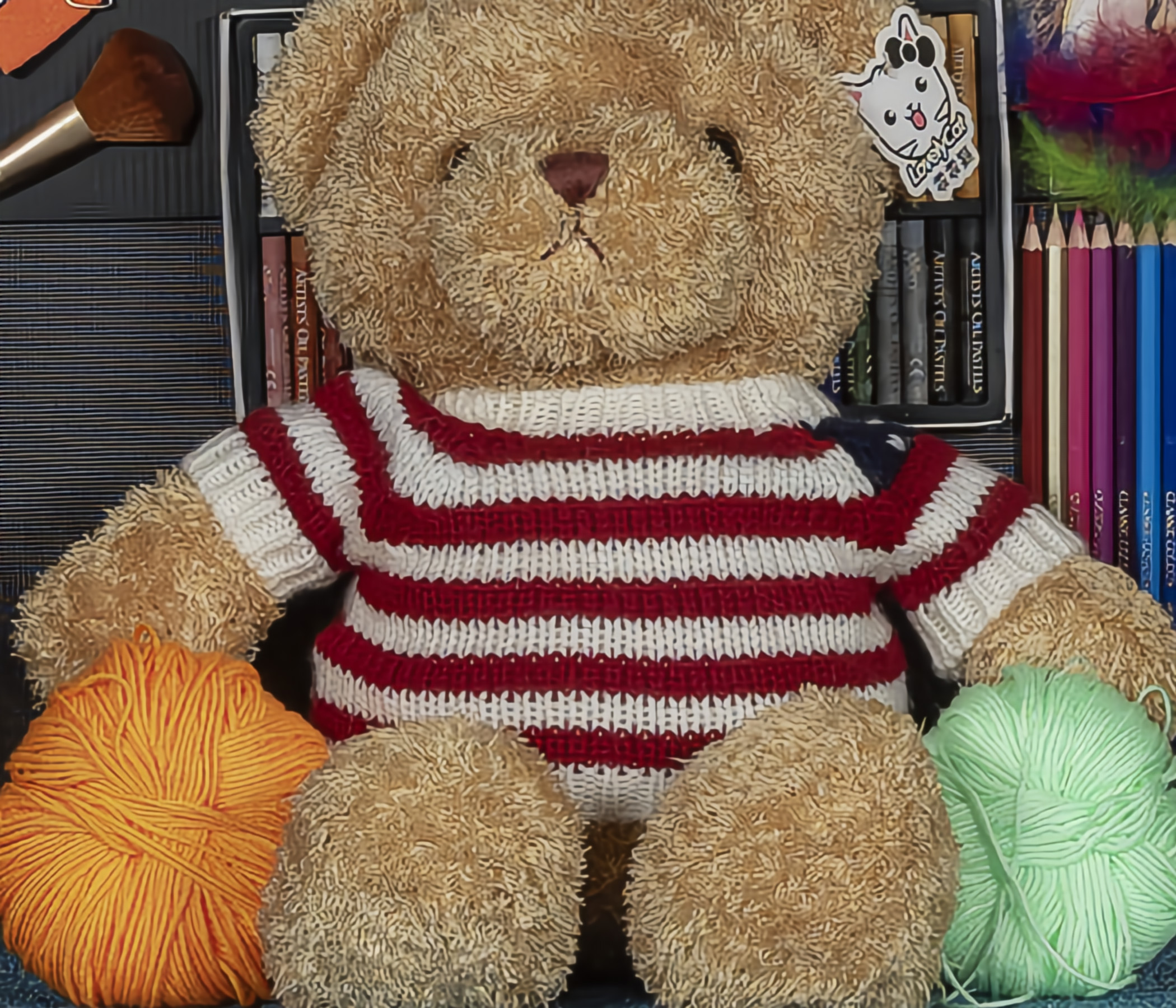}
    \vspace{0.3mm}
    \\
    \vspace{0.5mm}
    & 25.74 dB & 20.04 dB & 25.48 dB & 27.25 dB & 24.33 dB
    \\
    \hspace{-3mm}
    \multirow{6}{*}{\makecell{MIRNet \\ (Ours)}}
    &  \includegraphics[trim={1100 550 150 900 },clip,width=.177\textwidth,valign=t]{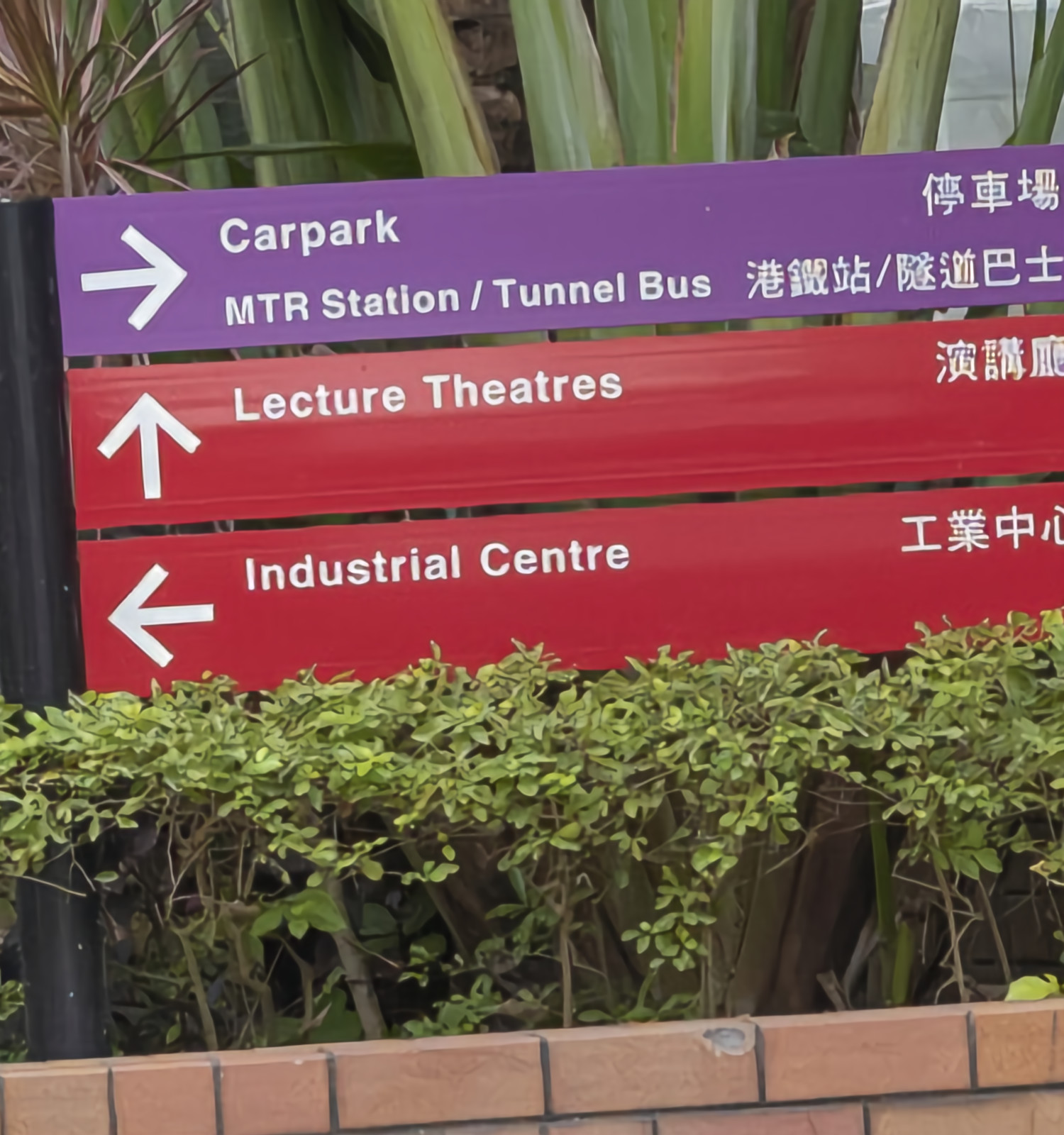}
    &  \includegraphics[trim={320 180 930 870 },clip,width=.177\textwidth,valign=t]{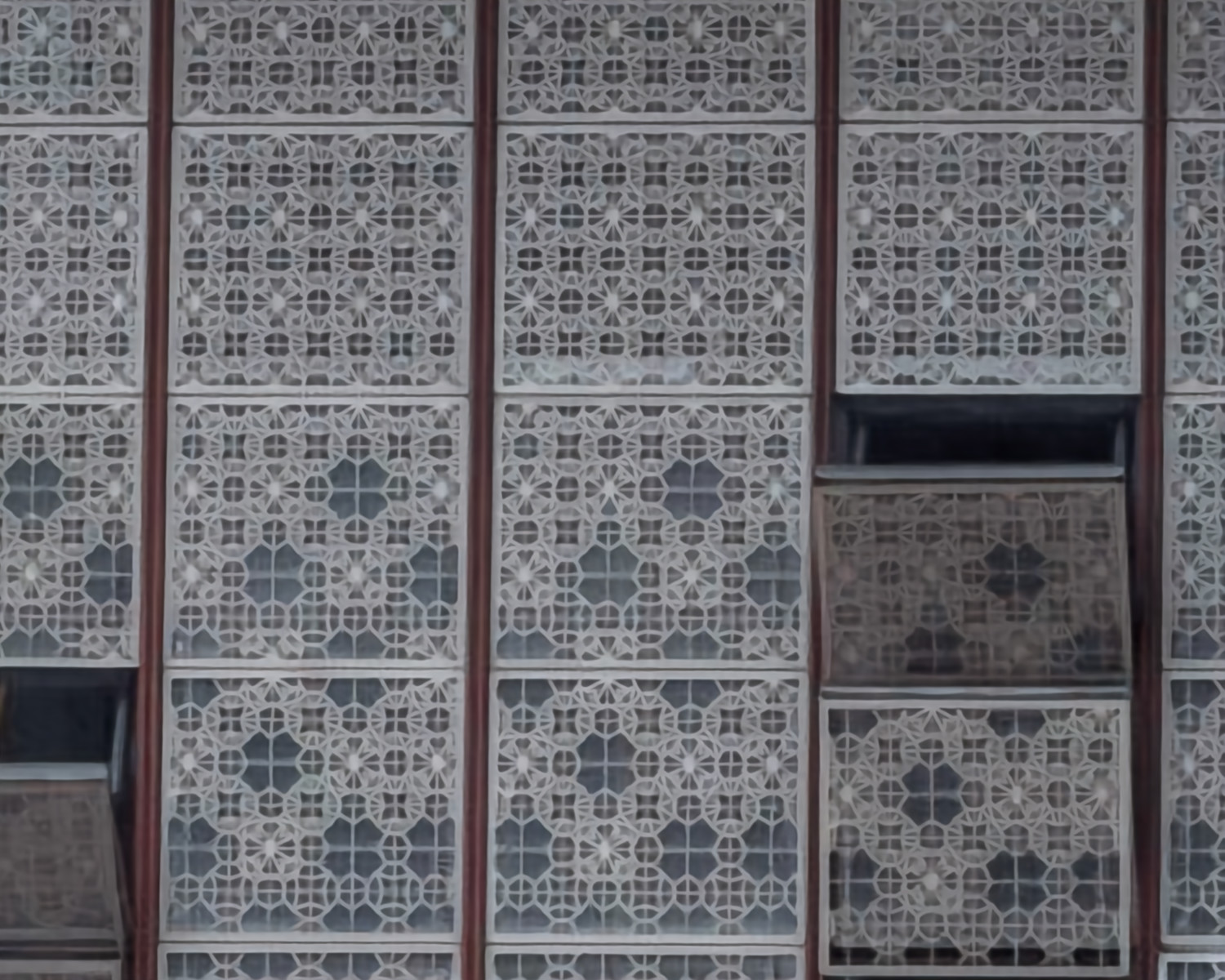}
    &  \includegraphics[trim={250 100 300 850 },clip,width=.177\textwidth,valign=t]{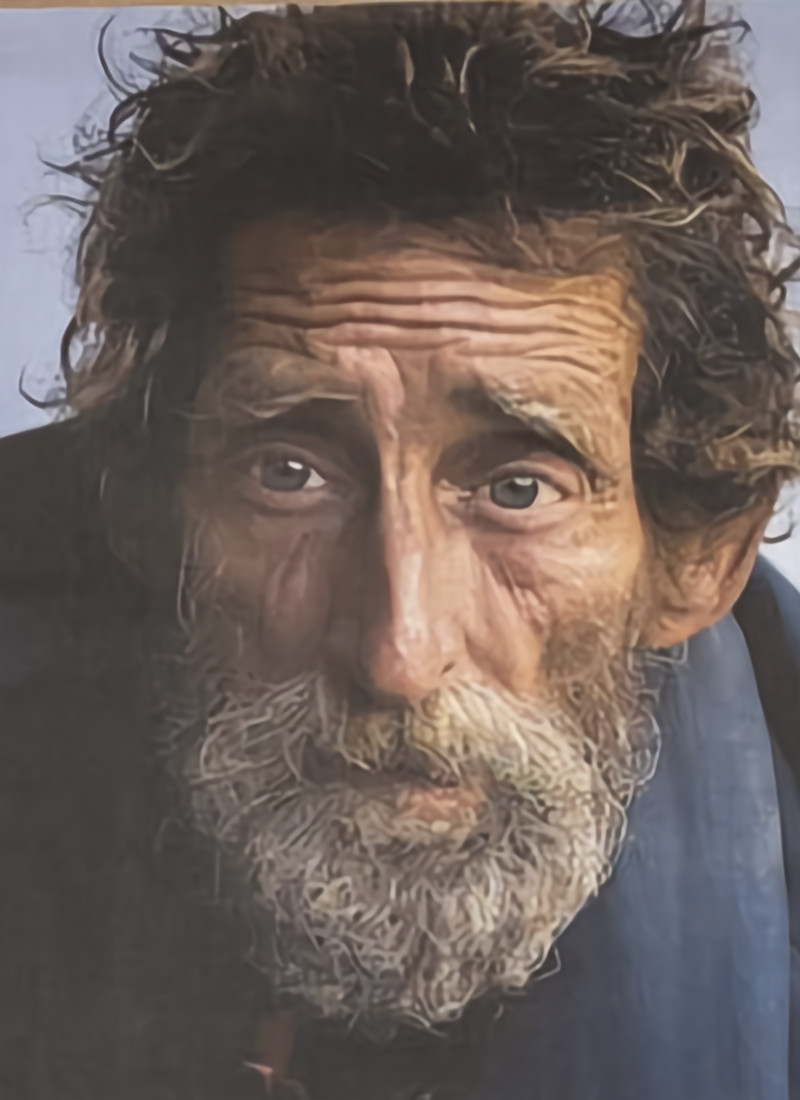}
    &  \includegraphics[trim={350 550 700 300 },clip,width=.177\textwidth,valign=t]{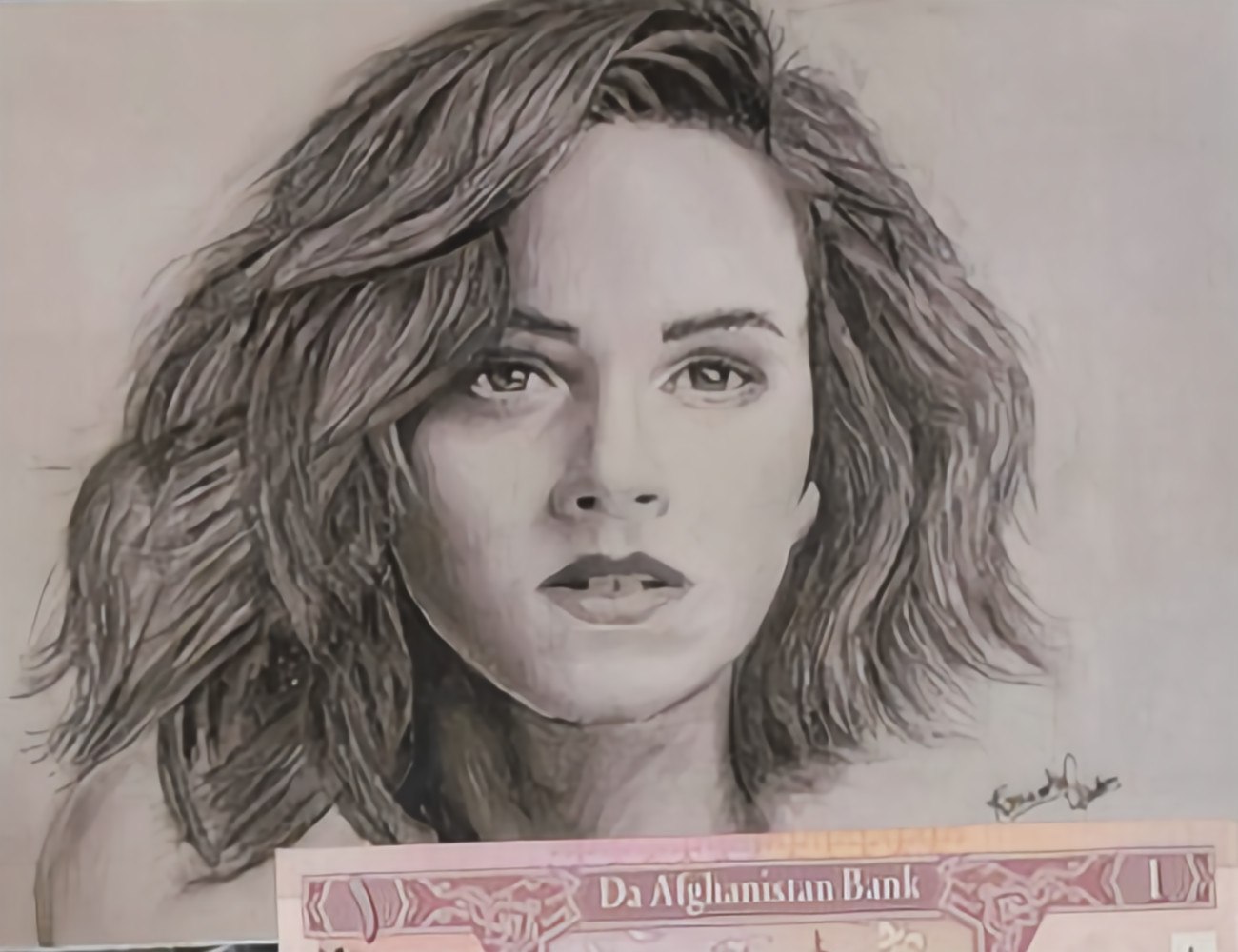}
    &  \includegraphics[trim={150 250 1700 1400 },clip,width=.177\textwidth,valign=t]{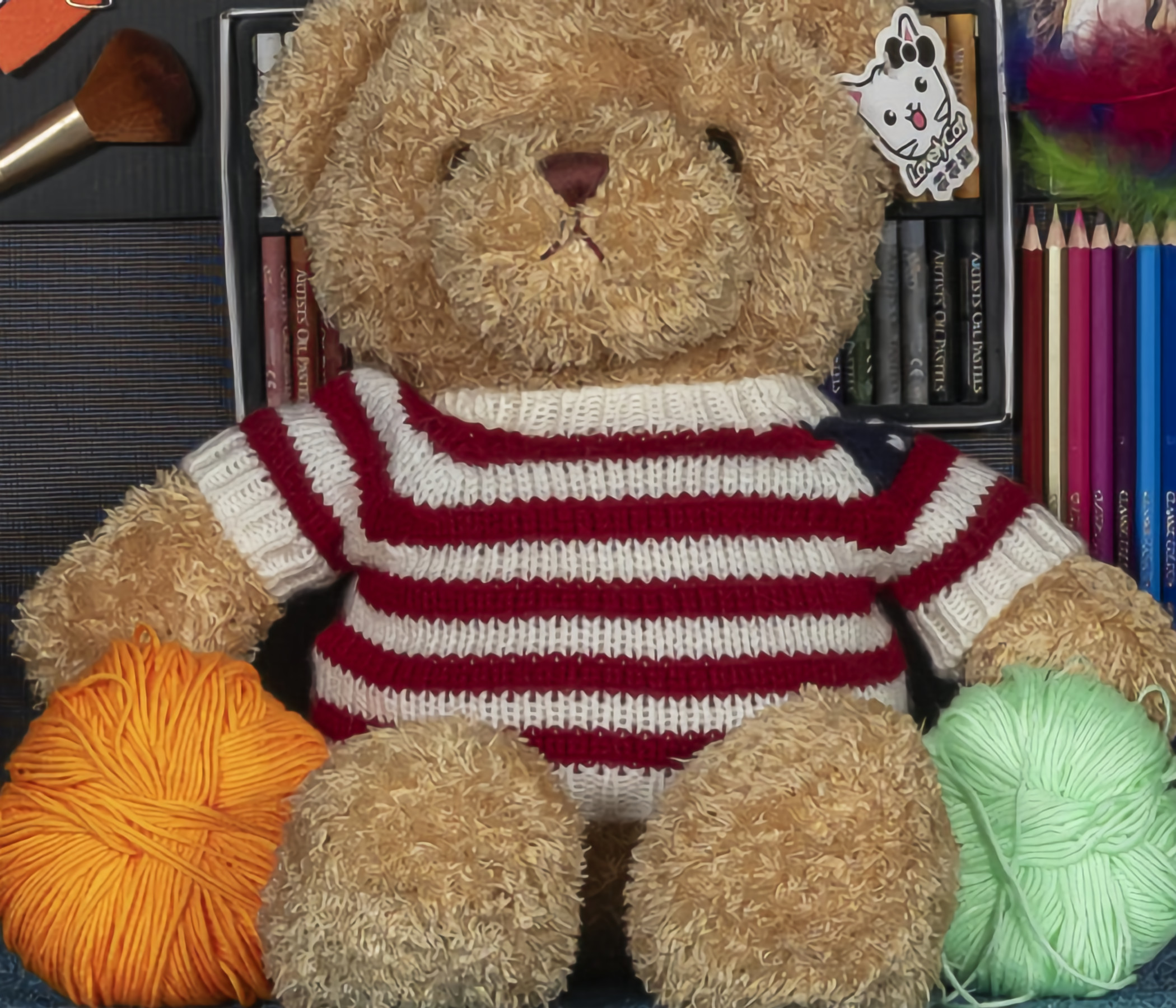}
    \vspace{0.3mm}
    \\
    & \textbf{27.22 dB} & \textbf{21.23 dB} & \textbf{27.04 dB} & \textbf{29.49 dB} & \textbf{26.87 dB}
\end{tabular}}
\end{center}
\vspace{-6mm}
\caption{\small Additional visual examples for $\times4$ super-resolution, comparing our MIRNet against the previous best approach~\cite{RealSR}. Note that all example crops are taken from different images. The full-resolution versions (and many more examples) are provided in the supplementary material. }
\label{fig:sr crop examples}
\vspace{-1.5em}
\end{figure}

\begin{table}[t]
\begin{center}
\caption{\small Cross-camera generalization test for super-resolution. Networks trained for one camera are tested on the other camera. Our MIRNet shows good generalization for all possible cases.}
\label{table:realSR generalization}
\setlength{\tabcolsep}{8.5pt}
\scalebox{0.70}{
\begin{tabular}{l | c | c | c c | c c | c c}
\toprule
\rowcolor{color3}  &  & Bicubic & \multicolumn{2}{c|}{RCAN~\cite{RCAN} (Trained on)} & \multicolumn{2}{c|}{LP-KPN~\cite{RealSR} (Trained on)} & \multicolumn{2}{c}{MIRNet (Trained on)}  \\
\cline{4-9}
\rowcolor{color3} Tested on & Scale &  & Canon & Nikon & Canon & Nikon & Canon & Nikon \\
\midrule
&        $\times$2 &   33.05   &  34.34  &  34.11 &   34.38  &  34.18   &  \textbf{35.41}  & \textbf{35.14} \\
Canon & $\times$3 &    29.67   &   30.65  & 30.28 &   30.69 & 30.33    &  \textbf{31.97}  & \textbf{31.56} \\
&        $\times$4 &  28.31   &  29.46  &  29.04 &   29.48  &   29.10   &  \textbf{30.35} & \textbf{29.95} \\ 
\midrule
&        $\times$2 &   31.66   &   32.01  &  32.30 &   32.05  &  32.33   &  \textbf{32.58}  & \textbf{ 33.19}  \\
Nikon &   $\times$3 &    28.63   &   29.30  &  29.75 &   29.34  &  29.78    &  \textbf{29.71}  &\textbf{ 30.05}  \\
&         $\times$4 &  27.28   &  27.98  &  28.12 &   28.01 &  28.13   &  \textbf{28.16}  & \textbf{28.37}  \\
\bottomrule
\end{tabular}}
\end{center}
\vspace{-2.6em}
\end{table}

\begin{table}[t]
\begin{center}
\caption{\small Low-light image enhancement evaluation on the LoL dataset \cite{wei2018deep}. The proposed method significantly advances the state-of-the-art.}
\label{table:lol}
\setlength{\tabcolsep}{4.5pt}
\scalebox{0.7}{
\begin{tabular}{l c c c c c c c c c c c c c}
\toprule
  \rowcolor{color3} Method & BIMEF  & CRM & Dong  & LIME   & MF  & RRM  & SRIE & Retinex-Net  & MSR  & NPE & GLAD  & KinD  & MIRNet \\
 \rowcolor{color3}  & \cite{ying2017bio} & \cite{ying2017new} & \cite{dong2011fast} &  \cite{guo2016lime}  &  \cite{fu2016weighted} &  \cite{liu2018structure} &  \cite{fu2016weighted} &  \cite{wei2018deep} &  \cite{jobson1997multiscale} &  \cite{wang2013naturalness} &  \cite{wang2018gladnet} &  \cite{zhang2019kindling}  & (Ours)  \\
   
 \midrule
PSNR & 13.86 & 17.20 & 16.72 & 16.76 & 18.79 & 13.88 & 11.86 & 16.77 & 13.17 & 16.97 & 19.72 & 20.87 & \textbf{24.14}\\
SSIM & 0.58 & 0.64 & 0.58 & 0.56 & 0.64 & 0.66 & 0.50 & 0.56 & 0.48 & 0.59 & 0.70 & 0.80 & \textbf{0.83}\\
\bottomrule
\end{tabular}}
\end{center}
\vspace{-2.6em}
\end{table}

\begin{table}[!t]
\begin{center}
\caption{\small Image enhancement comparisons on the MIT-Adobe FiveK dataset \cite{mit_fivek}.} 
\label{table:fivek}
\setlength{\tabcolsep}{11pt}
\scalebox{0.72}{
\begin{tabular}{l c c c c c c}
\toprule
 \rowcolor{color3} Method & HDRNet \cite{Gharbi2017} & W-Box \cite{hu2018exposure} & DR \cite{park2018distort} & DPE \cite{chen2018deep}  & DeepUPE \cite{wang2019underexposed}  & MIRNet (Ours) \\
 \midrule
PSNR & 21.96 & 18.57 & 20.97 & 22.15 & 23.04 & \textbf{23.73}\\
SSIM & 0.866 & 0.701 & 0.841 & 0.850 & 0.893 & \textbf{0.925}\\
\bottomrule
\end{tabular}}
\end{center}
\vspace{-2.2em}
\end{table}

Visual comparisons in Fig.~\ref{fig:sr example} show that our MIRNet recovers content structures effectively. In contrast, VDSR~\cite{VDSR}, SRResNet~\cite{SRResNet} and RCAN~\cite{RCAN} reproduce results with noticeable artifacts. 
Furthermore, LP-KPN~\cite{RealSR} is not able to preserve structures (see near the right edge of the crop). 
Several more examples are provided in Fig.~\ref{fig:sr crop examples} to further compare the image reproduction quality of our method against the previous best method~\cite{RealSR}.  
It can be seen that LP-KPN~\cite{RealSR} has a tendency to over-enhance the contrast (cols. 1, 3, 4) and in turn causes loss of details near dark and high-light areas.
In contrast, the proposed MIRNet successfully reconstructs structural patterns and edges (col. 2) and produces images that are natural (cols. 1, 4) and have better color reproduction (col. 5). 

\vspace{0.4em}\noindent\textbf{Cross-camera generalization.} The RealSR~\cite{RealSR} dataset consists of images taken with Canon and Nikon cameras at three scaling factors. To test the cross-camera generalizability of our method, we train the network on the training images of one camera and directly evaluate it on the test set of the other camera. Table~\ref{table:realSR generalization} demonstrates the generalization of  competing methods for four possible cases: (a) training and testing on Canon, (b) training on Canon, testing on Nikon, (c) training and testing on Nikon, and (d) training on Nikon, testing on Canon. It can be seen that, for all scales, LP-KPN~\cite{RealSR} and RCAN~\cite{RCAN} shows comparable performance. 
In contrast, our MIRNet exhibits more promising generalization.


\subsection{Image Enhancement}

In this section, we demonstrate the effectiveness of our algorithm by evaluating it for the image enhancement task. 
We report PSNR/SSIM values of our method and several other techniques in Table~\ref{table:lol} and Table~\ref{table:fivek} for the LoL~\cite{wei2018deep} and MIT-Adobe FiveK~\cite{mit_fivek} datasets, respectively. 
It can be seen that our MIRNet achieves significant improvements over previous approaches. 
Notably, when compared to the recent best methods, MIRNet obtains $3.27$~dB performance gain over KinD~\cite{zhang2019kindling} on the LoL dataset and $0.69$~dB improvement over DeepUPE\footnote{Note that the quantitative results reported in~\cite{wang2019underexposed} are incorrect. The correct scores are later released by the original authors~\href{https://drive.google.com/file/d/1fJ7MQfm6NuCMtfQzLM0Y6LNU9XyQb6Ho/view}{[link]}.} \cite{wang2019underexposed} on the Adobe-Fivek dataset.  

\begin{figure}[!t]
  \begin{center}
    \tiny
    \begin{tabular}{cccc}
      \includegraphics[width=0.244\textwidth]{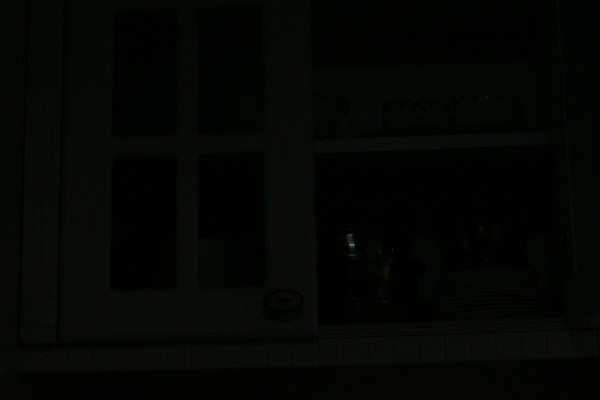}&\hspace{-1.5mm}
      \includegraphics[width=0.244\textwidth]{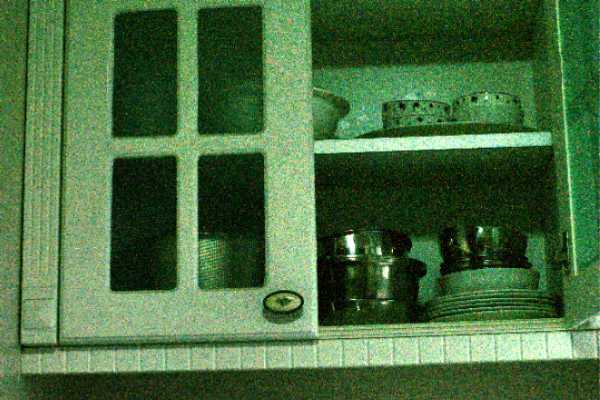}&\hspace{-1.5mm}
       \includegraphics[width=0.244\textwidth]{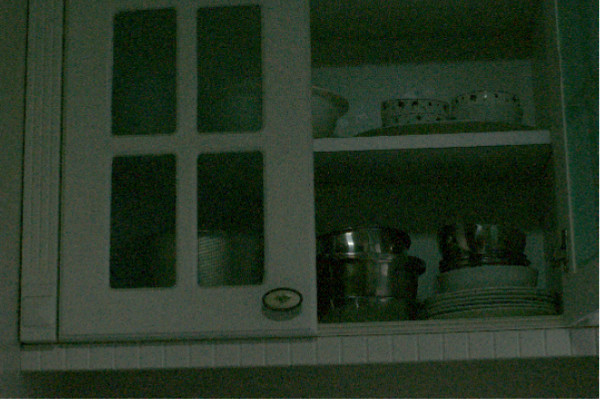}&\hspace{-1.5mm}
       \includegraphics[width=0.244\textwidth]{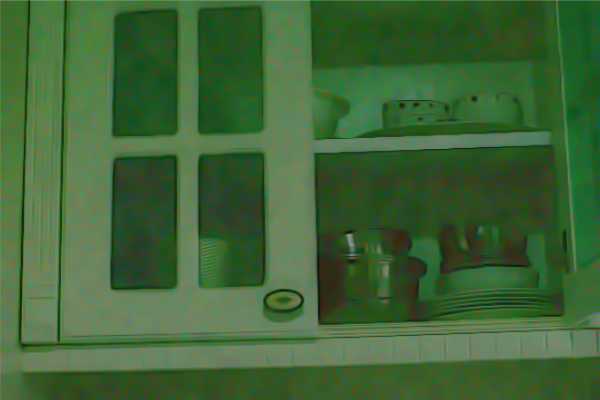}\hspace{-1.5mm}
       \vspace{0.3mm}
       \\
       \vspace{0.5mm}
      Input image & LIME \cite{guo2016lime} & CRM~\cite{ying2017bio} & Retinex-Net \cite{wei2018deep} \\
      \includegraphics[width=0.244\textwidth]{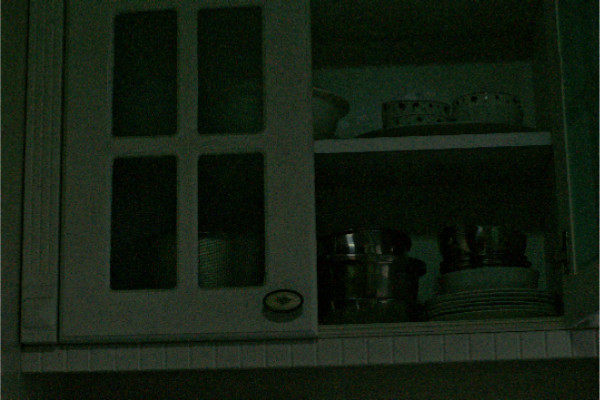}&\hspace{-1.5mm}
      \includegraphics[width=0.244\textwidth]{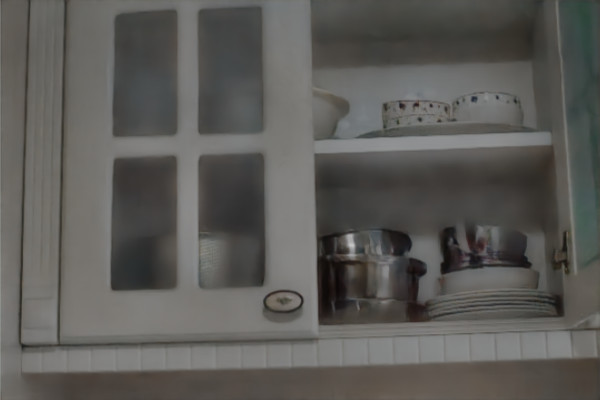}&\hspace{-1.5mm}
      \includegraphics[width=0.244\textwidth]{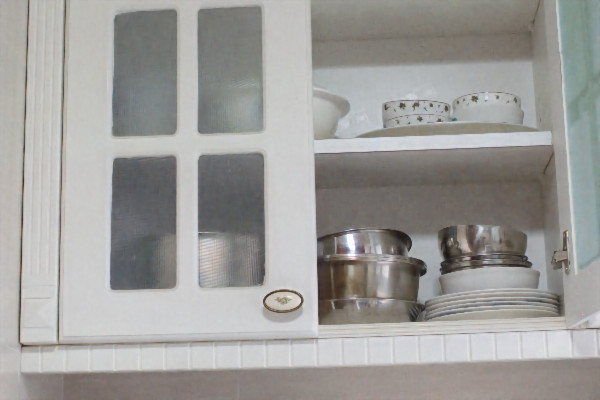}&\hspace{-1.5mm}
      \includegraphics[width=0.244\textwidth]{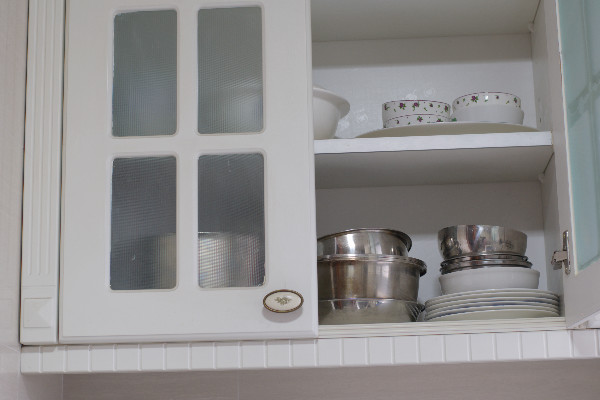}\hspace{-1.5mm}
      \vspace{0.3mm}
      \\
      \vspace{0.5mm}
       SRIE \cite{fu2016weighted} &  KinD \cite{zhang2019kindling} & MIRNet (Ours) & Ground-truth
    \end{tabular}
  \end{center}\vspace{-2em}
    \caption{\small Visual comparison of low-light enhancement approaches on the LoL dataset~\cite{wei2018deep}. Our method reproduces image that is visually closer to the ground-truth in terms of brightness and global contrast.}
    \label{Fig:qual_lol}
\end{figure}

\begin{figure}[!t]
  \begin{center}
    \tiny
    \begin{tabular}{ccc}
      \includegraphics[width=0.324\textwidth]{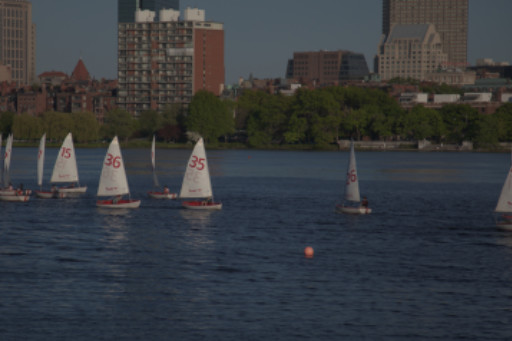}&\hspace{-1.5mm}
      \includegraphics[width=0.324\textwidth]{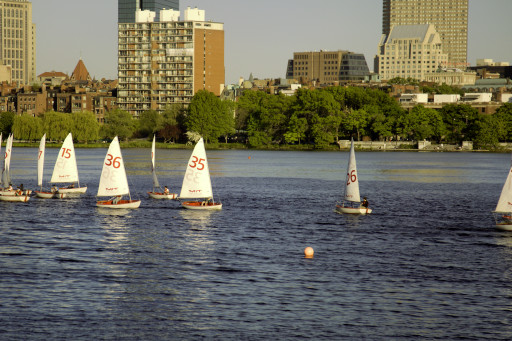}&\hspace{-1.5mm}
      \includegraphics[width=0.324\textwidth]{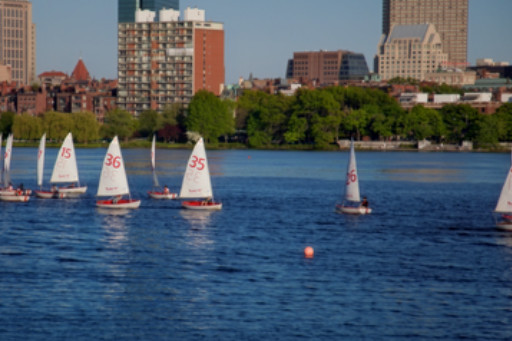}\hspace{-1.5mm}
      \vspace{0.3mm}
      \\
      \vspace{0.5mm}
      Input image & HDRNet~\cite{Gharbi2017} & DPE~\cite{chen2018deep} \\
      \includegraphics[width=0.324\textwidth]{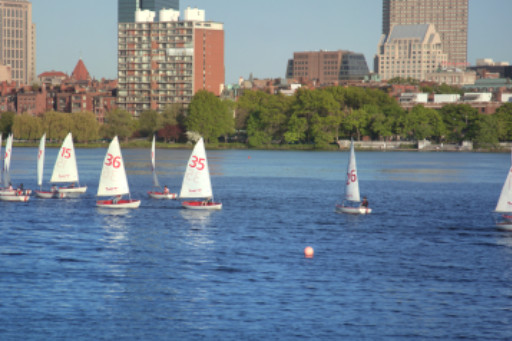}&\hspace{-1.5mm}
      \includegraphics[width=0.324\textwidth]{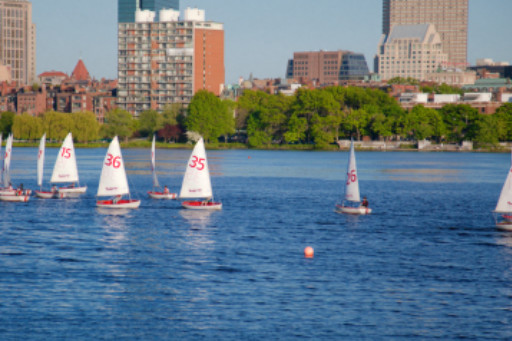}&\hspace{-1.5mm}
      \includegraphics[width=0.324\textwidth]{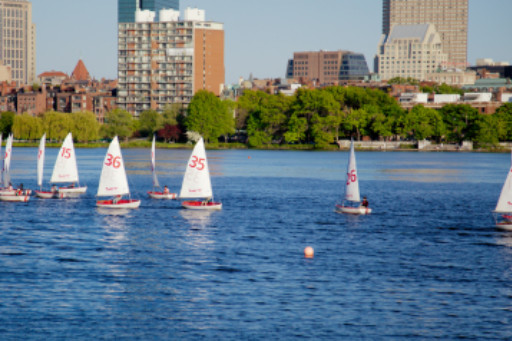}\hspace{-1.5mm}
      \vspace{0.3mm}
      \\
      \vspace{0.5mm}
       DeepUPE \cite{wei2018deep} & MIRNet (Ours)  & Ground-truth

    \end{tabular}
  \end{center}\vspace{-2em}
    \caption{\small Visual results of image enhancement on the MIT-Adobe FiveK~\cite{mit_fivek} dataset. 
    Compared to the state-of-the-art, our MIRNet makes better color and contrast adjustments and produces image that is vivid, natural and pleasant in appearance. }
      \vspace{-1em}
    \label{Fig:qual_fivek}\vspace{-0em}
\end{figure}

We show visual results in Fig.~\ref{Fig:qual_lol} and Fig.~\ref{Fig:qual_fivek}. Compared to other techniques, our method generates enhanced images that are natural and vivid in appearance and have better global and local contrast.

\section{Ablation Studies}
We study the impact of each of our architectural components and design choices on the final performance. All the ablation experiments are performed for the super-resolution task with $\times3$ scale factor. 
Table~\ref{table:ablation main} shows that removing skip connections causes the largest performance drop. Without skip connections, the network finds it difficult to converge and yields high training errors, and consequently low PSNR.
Furthermore, the information exchange among parallel convolution streams via SKFF is helpful and leads to improved performance. 
Similarly, DAU also makes a positive influence to the final image quality. 

Next, we analyze the feature aggregation strategy in Table~\ref{table:ablation aggregation}. It shows that the proposed SKFF generates favorable results compared to summation and concatenation. Moreover, it can be seen that our SKFF uses $\sim6\times$ fewer parameters than concatenation. 
Finally, in Table~\ref{table: ablation MRB} we study how the number of convolutional streams and columns (DAU blocks) of MRB affect the image restoration quality. 
We note that increasing the number of streams provides significant improvements, thereby justifying the importance of multi-scale features processing. Moreover, increasing the number of columns yields better scores, thus indicating the significance of information exchange among parallel streams for feature consolidation.   
Additional ablation studies and qualitative results are provided in the supplementary material.

\begin{table}[t]
\parbox{.45\linewidth}{
\centering
\caption{\small Impact of individual components of MRB.}
\label{table:ablation main}
\setlength{\tabcolsep}{2pt}
\scalebox{0.70}{
\begin{tabular}{l c c c c c}
\toprule
Skip connections   &      & \ch  &\ch   & \ch  &    \ch  \\ 
DAU                & \ch  &      &\ch   &      &    \ch  \\  
SKFF intermediate  & \ch  & \ch  &      &      &    \ch \\
SKFF final         & \ch  & \ch  &\ch   &\ch   &    \ch \\
\midrule
\rowcolor{color3} PSNR (in dB)   & 27.91 & 30.97 & 30.78 & 30.57 & \textbf{31.16} \\  \bottomrule
\end{tabular}}
}
\hfill
\parbox{.48\linewidth}{
\centering
\caption{\small Feature aggregation. Our SKFF uses $\sim6\times$ fewer parameters than concat, but generates better results. }
\label{table:ablation aggregation}
\setlength{\tabcolsep}{12.5pt}
\scalebox{0.70}{
\begin{tabular}{l c c c }
\toprule
\rowcolor{color3} Method    & Sum     & Concat  & SKFF  \\ 
\midrule
PSNR (in dB) & 30.76 & 30.89  & \textbf{31.16} \\
Parameters   & 0     & 12,288 & 2,049          \\  \bottomrule
\end{tabular}}
}
\vspace*{-1.5em}
\end{table}

\begin{table}[t]
\begin{center}
\caption{\small Ablation study on different layouts of MRB. \emph{Rows} denote the number of parallel resolution streams, and \emph{Cols} represent the number of columns containing DAUs. }
\label{table: ablation MRB}
\setlength{\tabcolsep}{7pt}
\scalebox{0.70}{
\begin{tabular}{l | c c c | c c c | c c c }
\toprule
\rowcolor{color3} & \multicolumn{3}{c|}{Rows = 1} & \multicolumn{3}{c|}{Rows = 2} & \multicolumn{3}{c}{Rows = 3} \\
\cline{2-10}
\rowcolor{color3}  & Cols = 1 & Cols = 2 & Cols = 3 & Cols = 1 & Cols = 2 & Cols = 3 & Cols = 1 & Cols = 2 & Cols = 3 \\
\midrule
PSNR &   29.92   &  30.11  &   30.17  &  30.15 &   30.83  &  30.92   &  30.24  & 31.16 &  31.18  \\
\bottomrule
\end{tabular}}
\end{center}\vspace{-2.5em}
\end{table}

\section{Concluding Remarks}

Conventional image restoration and enhancement pipelines either stick to the full resolution features along the network hierarchy or use an encoder-decoder architecture. The first approach helps retain precise spatial details, while the latter one provides better contextualized representations. However, these methods can satisfy only one of the above two requirements, although real-world image restoration tasks demand a combination of both conditioned on the given input sample. In this work, we propose a novel architecture whose main branch is dedicated to full-resolution processing and the complementary set of parallel branches provides better contextualized features. We propose novel mechanisms to learn relationships between features within each branch as well as across multi-scale branches. Our feature fusion strategy ensures that the receptive field can be dynamically adapted without sacrificing the original feature details. Consistent achievement of state-of-the-art results on five datasets for three image restoration and enhancement tasks corroborates the effectiveness of our approach. 

\vspace{0.5em}\noindent\textbf{Acknowledgments.} Ming-Hsuan Yang is supported by the NSF CAREER Grant 1149783.
%
%
\bibliographystyle{splncs04}
\bibliography{MIRNet}
\end{document}